\documentclass[10pt,twocolumn,letterpaper]{article}

\usepackage[pagenumbers]{cvpr} 

\usepackage{graphicx}
\usepackage{amsmath}
\usepackage{amssymb}
\usepackage{booktabs}
\usepackage{bm}
\usepackage{multirow}
\usepackage{makecell}
\usepackage{wrapfig}

\usepackage[pagebackref,breaklinks,colorlinks]{hyperref}

\usepackage[capitalize]{cleveref}
\crefname{section}{Sec.}{Secs.}
\Crefname{section}{Section}{Sections}
\Crefname{table}{Table}{Tables}
\crefname{table}{Tab.}{Tabs.}

\def\cvprPaperID{8485} 
\def\confName{CVPR}
\def\confYear{2023}

\begin{document}

\title{Learning Visibility Field for Detailed 3D Human Reconstruction and Relighting}

\author{Ruichen Zheng$^{*,1,2}$,    Peng Li$^{*,1}$,    Haoqian Wang$^{1}$,    Tao Yu$^{1}$\\
$^1$Tsinghua University, China       $^2$Weilan Tech, Beijing, China
}
\maketitle
\begin{abstract}
Detailed 3D reconstruction and photo-realistic relighting of digital humans are essential for various applications. To this end, we propose a novel sparse-view 3d human reconstruction framework that closely incorporates the occupancy field and albedo field with an additional visibility field--it not only resolves occlusion ambiguity in multiview feature aggregation, but can also be used to evaluate light attenuation for self-shadowed relighting. To enhance its training viability and efficiency, we discretize visibility onto a fixed set of sample directions and supply it with coupled geometric 3D depth feature and local 2D image feature. We further propose a novel rendering-inspired loss, namely TransferLoss, to implicitly enforce the alignment between visibility and occupancy field, enabling end-to-end joint training. Results and extensive experiments demonstrate the effectiveness of the proposed method, as it surpasses state-of-the-art in terms of reconstruction accuracy while achieving comparably accurate relighting to ray-traced ground truth.

\end{abstract}

\section{Introduction}
\label{sec:intro}

3D reconstruction and relighting are of great importance in human digitization, especially in supporting realistic rendering in varying virtual environments, that can be widely applied in AR/VR \cite{ma2021pixel, orts2016holoportation}, holographic communication \cite{lawrence2021project, zhang2022virtualcube}, movie and gaming industry \cite{debevec2000pursuing}. 

Traditional methods often require dense camera setups using multi-view stereo, non-rigid registration and texture mapping \cite{fanello2014learning, guo2015robust}. To enhance capture realism, researchers have extended them with additional synchronous variable illumination systems, which aid photometric stereo for detail reconstruction and material acquisition \cite{guo2019relightables}. However, these systems are often too complex, expensive and difficult to maintain, thus preventing widespread applications.

By leveraging deep prior and neural representation, sophisticated dense camera setups can be reduced to a single camera, leading to blossoms in learning-based human reconstruction. In particular, encoding human geometry and appearance as continuous fields using Multi-Layer Perceptron (MLP) has emerged as a promising lead. Starting from Siclope \cite{natsume2019siclope} and PIFu \cite{saito2019pifu}, a series of methods \cite{hong2021stereopifu} improve the reconstruction performance in speed \cite{li2020monocular, feng2022fof}, quality \cite{saito2020pifuhd}, robustness \cite{zheng2021pamir, zheng2019deephuman} and light decoupling \cite{alldieck2022photorealistic}. However, single-view reconstruction quality is restricted by its inherent depth ambiguity, thus limiting its application under view-consistent high-quality requirements.

Therefore, as the trade-off between view coverage and system accessibility, sparse-view reconstruction has become a research hotpot. The predominant practice is to project the query point onto each view to interpolate local features, which are then aggregated and fed to MLP for inference \cite{saito2019pifu, yu2021function4d, suo2021neuralhumanfvv, shao2022diffustereo, chibane2020implicit, huang2021dynamic}. This method suffers from occlusion ambiguity, where some views may well be occluded, and mixing their features with visible ones causes inefficient feature utilization, thus penalizing the reconstruction quality \cite{saito2019pifu}. A natural solution is to filter features based on view visibility. Human templates such as SMPL \cite{loper2015smpl} can serve as effective guidance \cite{bhatnagar2020ipnet, zheng2021deepmulticap, peng2021neural, verbin2022ref}, but introduce additional template alignment errors and therefore cannot guarantee complete occlusion awareness. Function4D \cite{yu2021function4d} leverages the truncated Projective Signed Distance Function (PSDF) for visibility indication, but its level of details is susceptible to depth noise.

To this end, we directly model a continuous visibility field, which can be efficiently learned with our proposed framework and discretization technique using sparse-view RGB-D input. The visibility field enables efficient visibility query, which effectively guides multi-view feature aggregation for more accurate occupancy and albedo inference. Moreover, visibility can also be directly used for light attenuation evaluation--the key ingredient in achieving realistic self-shadowing. When supervising jointly with our novel TransferLoss, the alignment between the visibility field and occupancy field can be implicitly enforced without between-field constraints, such as matching visibility with occupancy ray integral. We train our framework end-to-end and demonstrate its effectiveness in detailed 3D human reconstruction by quantity and quality comparison with the state-of-the-art. We directly relight the reconstructed geometry with inferred visibility using diffuse Bidirectional Reflectance Distribution Function (BRDF) as in \cref{fig_teaser}, which achieves photo-realistic self-shadowing without any post ray-tracing steps. To conclude, our contributions include:
\begin{itemize}
    \item An end-to-end framework for sparse-view detailed 3D human reconstruction that also supports direct self-shadowed relighting. 
    \item A novel method of visibility field learning, with the specifically designed TransferLoss significantly improves field alignment.
    \item A visibility-guided multi-view feature aggregation strategy that guarantees occlusion awareness.
\end{itemize}

\begin{figure*}
    \centering
    \includegraphics[width=\linewidth]{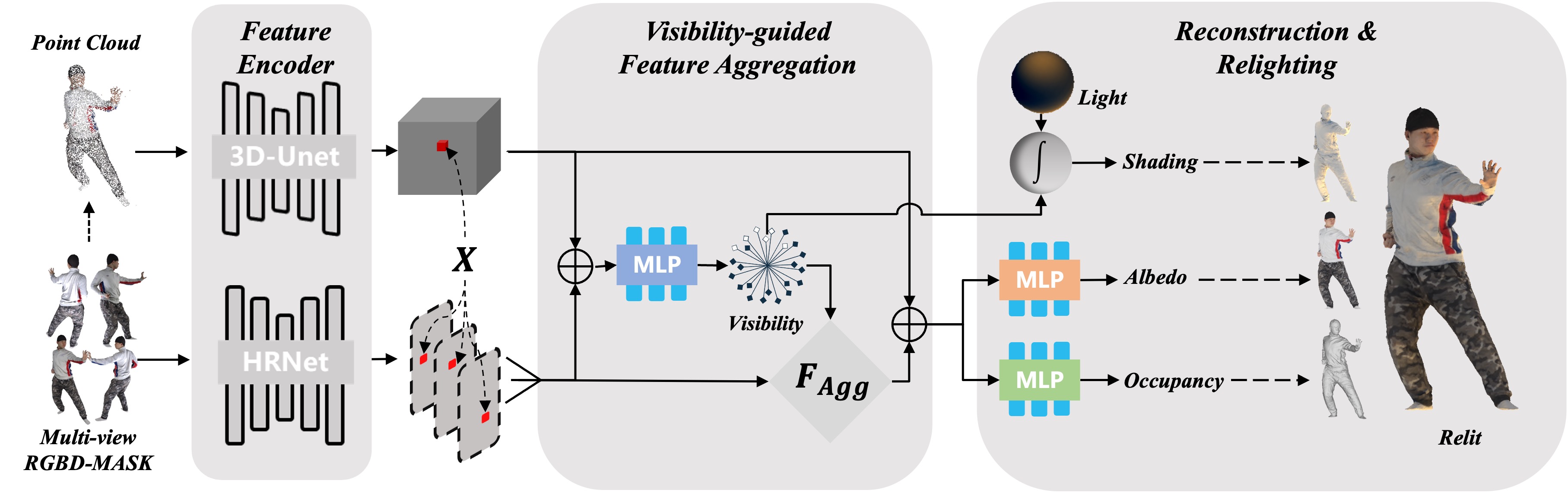}
    \caption{Method overview. Given sparse RGB-D frames as input, our framework first infer visibility, which is then applied to guide multi-view 2D feature aggregation for joint occupancy and albedo inference. Our framework is end-to-end trainable and produces high-fidelity human reconstruction that supports  direct self-shadowed relighting without using any post ray-tracing steps.}
    \label{fig:framework}
\end{figure*}

\section{Background and Related Work}
\label{sec:works}


\textbf{Neural human synthesis}
The literature on human reconstruction is vast and rapidly growing. Here, we only review and contrast with closely related work and refer readers to surveys \cite{tian2022recovering, xie2022neural} for comprehensive reviews.

Neural implicit representation encodes geometry as a function of spatial coordinate using MLP. This representation is appealing as it is naturally differentiable, has exceptional expressiveness yet maintains compact memory footprint. The pioneer work follows an encoder-decoder-like architecture, where the globally encoded feature is applied to condition spatial coordinates to infer low-level geometric details \cite{park2019deepsdf, mescheder2019occupancy, chen2019learning, michalkiewicz2019deep}. This highly unbalanced information flow limits its capability to represent mere simple shapes \cite{peng2020convolutional}.
PIFu \cite{saito2019pifu} proposes to replace the global feature with pixel-aligned local features, which captures convolutional inductive bias and achieves highly detailed human reconstruction. It inspires a variety of works, ranging from quality improvement \cite{saito2020pifuhd}, parametric model extension \cite{zheng2021pamir, bhatnagar2020ipnet}, animation support~\cite{he2021arch++, huang2020arch}, light estimation and relighting \cite{alldieck2022photorealistic}.

By averaging local features across views, the pixel-aligned framework can be easily extended to multi-view settings \cite{saito2019pifu}. However, simple averaging diminishes high frequency details, yielding overly smoothed geometry. Moreover, it treats all views as equally visible even for partially occluded regions, resulting in inaccurate and erroneous reconstruction.
Zhang \etal~\cite{zheng2021deepmulticap} addresses the occlusion ambiguity by incorporating the attention mechanism, which weights features by their learned cross-view correlations. Despite its promising performance, self-attention introduces substantial memory and computation overhead. Yu \etal~ \cite{yu2021function4d} additionally leverages global depth information, namely PSDF, to annihilate the ambiguity but is sensitive to depth noise. In contrast, we use robustly learned visibility to weight per-view contribution, which handles occlusion in a physically plausible manner while being more memory efficient to compute.


\textbf{Relighting} boils down to substituting the incident environment radiance and re-evaluating the rendering equation \cite{kajiya1986rendering}. For a scene of reflectors, the outgoing radiance reflected at the surface point $\bm{x}$ in direction $\bm{\omega}_o$ can be described as:
\begin{equation}
\label{eq:rendering_equation}
    L(\bm{x}, \bm{\omega}_o) = \int_{\Omega^+} L(\bm{x}, \bm{\omega}_i) \rho(\bm{x}, \bm{\omega}_i, \bm{\omega}_o) V(\bm{x}, \bm{\omega}_i) (\bm{n} \cdot \bm{\omega}_i) d \bm{\omega}_i
\end{equation}
where $\bm{n}$ is unit surface normal at $\bm{x}$, $\Omega^+$ is hemisphere of possible directions $\bm{\omega}_i$, $L(\bm{x}, \bm{\omega}_i)$ is incident radiance arriving $\bm{x}$ along $\bm{\omega}_i$, directly from environment or indirectly reflected. $\rho$ is the BRDF that models the surface reflectance. $V$ is the visibility function that describes whether light $\bm{x}$ is attenuated along $\bm{\omega}_i$, As the main contributor to realistic self-shadowing, visibility evaluation requires ray tracing geometry over all sample directions $\bm{\omega}_i$ per fragment \cite{ji2022geometry}, which is expensive and usually omitted \cite{sengupta2018sfsnet} or baked offline using Precomputed Radiance Transfer (PRT) \cite{sloan2002precomputed}. PRT rewrites \cref{eq:rendering_equation} with the following transfer function
\begin{equation}
\label{eq:transfer_function}
    T(\bm{x}, \bm{\omega}_i, \bm{\omega}_o) = \rho(\bm{x}, \bm{\omega}_i, \bm{\omega}_o) V(\bm{x}, \bm{\omega}_i) (\bm{n} \cdot \bm{\omega}_i)
\end{equation}
which is independent of light and can be precomputed and projected on the Spherical Harmonics (SH) basis as coefficients \cite{ramamoorthi2001efficient} ahead of time to save rendering cost.

However, PRT does not ameliorate visibility calculation complexity and requires recompute after deformation. Its limitation has been addressed by several learning-based approaches\cite{kanamori2019relighting, tajima2021relightingwild, rainer2022neural, lagunas2021single, li2019deep}, which predict transfer coefficients using deep neural networks. Instead of its SH parameterization, we predict raw visibility, as it (1) can be used for feature aggregation, and (2) is also well defined in free space that can be learned without surface evaluation.

NeRF \cite{mildenhall2021nerf} models the surface density and employs volume rendering for high fidelity novel view synthesis. It has been extended with reflectance \cite{boss2021nerd} and transfer function \cite{lyu2022neural} to support relighting. Visibility is naturally defined in the density field, but its evaluation requires the integration of multiple density queries along the ray, which is equally inefficient and needs to be accelerated, such as using occlusion map \cite{chen2022relighting4d} or MLP \cite{srinivasan2021nerv}. Although sharing similar ends, we achieve this with different means: (1) We emphasize accuracy by supervising with ground truth and use the TransferLoss to regularize the field alignment, rather than attempting to directly align the fields by matching inferred visibility with accumulated transmittance. (2) We extensively train our model on human scan dataset to ensure fast inference, rather than fitting it per-scene in order to render arbitrary within-scene objects.

\section{Method}
\label{sec:method}
To introduce our method, we first define the visibility field and its discretization process, which prioritizes efficiency without compromising performance. We then outline our framework and visibility learning procedure. Finally, we showcase its application in sparse-view 3D human reconstruction, where it guides feature aggregation and enables direct self-shadowed relighting.


\begin{figure*}
\centering
  \begin{subfigure}{0.1\linewidth}
  \centering
    \raisebox{1pt}{
    \includegraphics[scale=0.3]{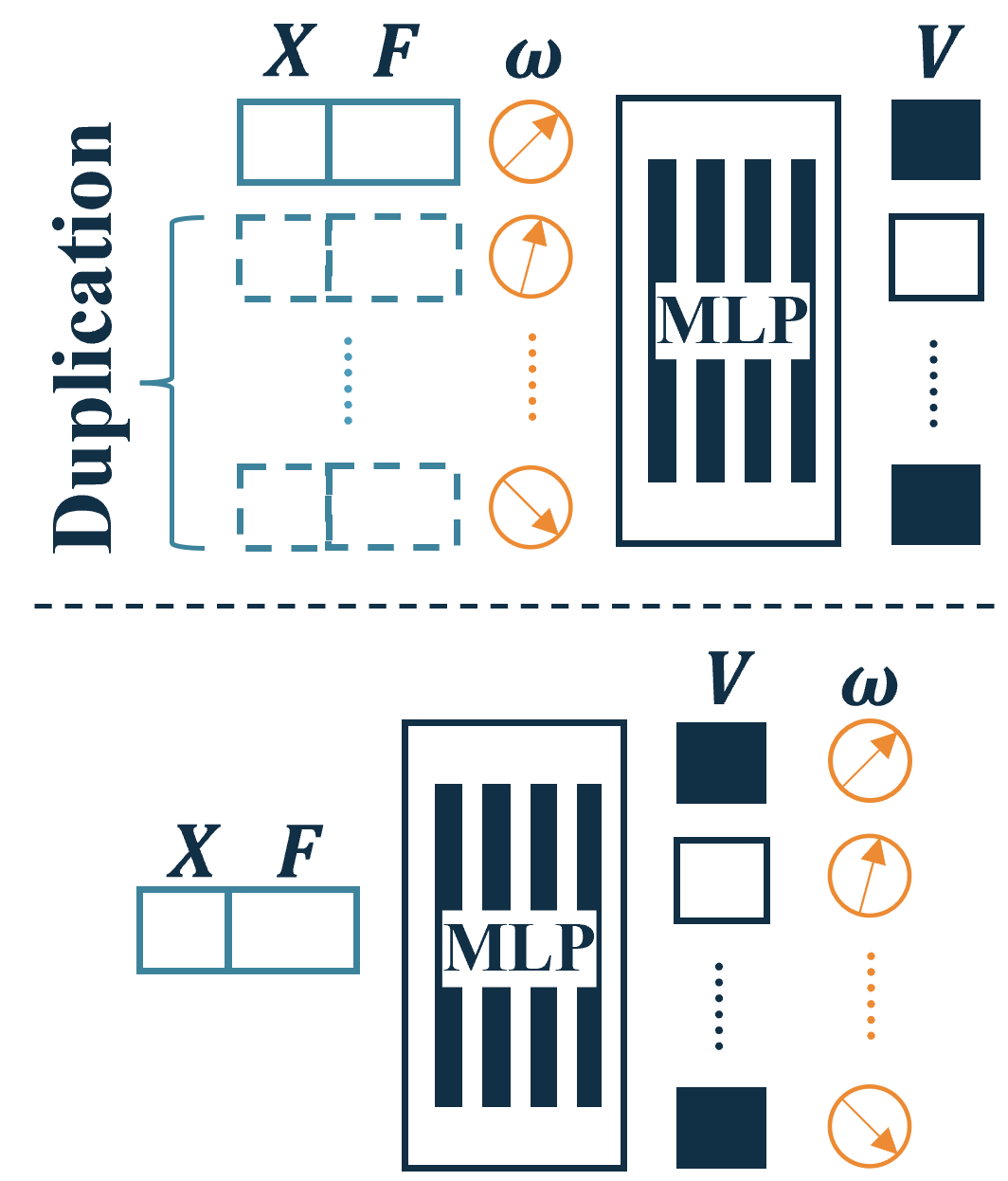}
    }
    \caption{}
    \label{fig:visibility_query}
  \end{subfigure}
     \hspace{1cm}
    \begin{subfigure}{0.2\linewidth}
     \centering
     \includegraphics[scale=0.4]{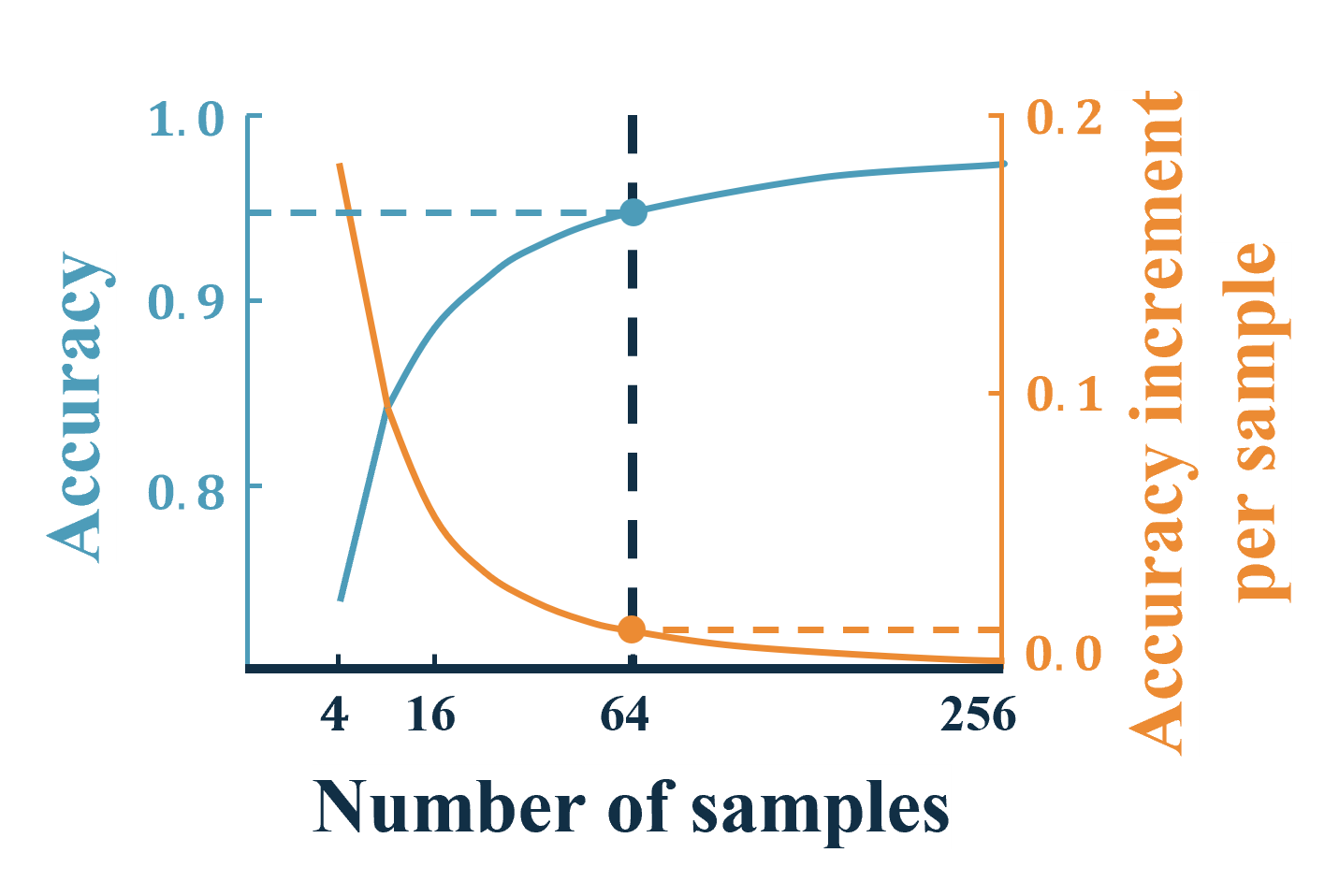}
     \caption{}
     \label{fig:sample_curve}
     \end{subfigure}
    \hspace{1cm}
    \begin{subfigure}{0.15\linewidth}
    \centering
    \raisebox{18pt}{\includegraphics[scale=0.45]{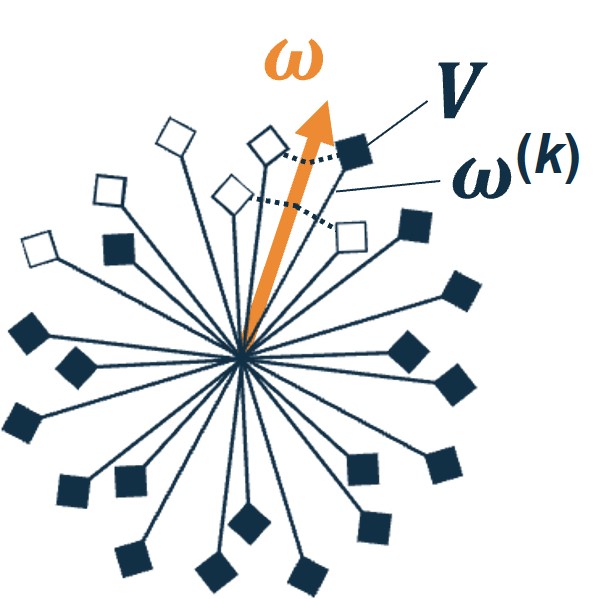}}
    \caption{}
    \label{fig:interpolation_fig}
    \end{subfigure}
    \hspace{-0.4cm}
    \begin{subfigure}{0.2\linewidth}
      \centering\includegraphics[scale=0.42]{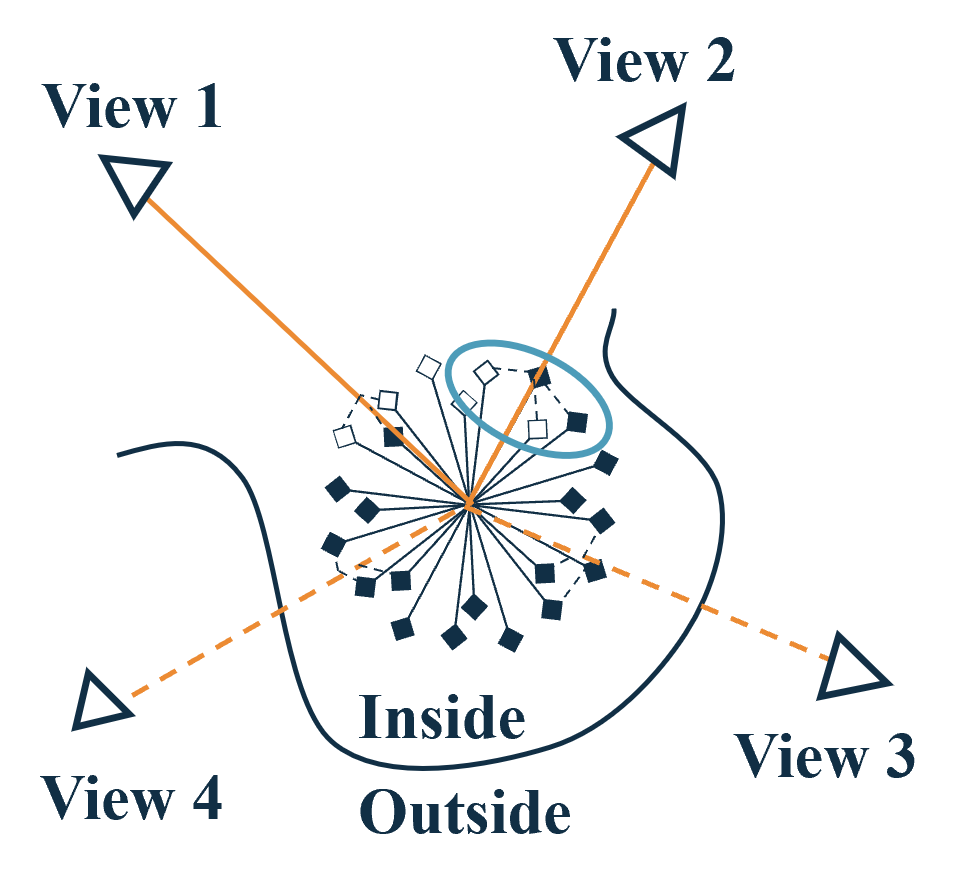}
      \caption{}
      \label{fig:aggregation}
     \end{subfigure}
     \hspace{-0.1cm}
     \begin{subfigure}{0.1\linewidth}
  \centering
    \includegraphics[scale=0.4]{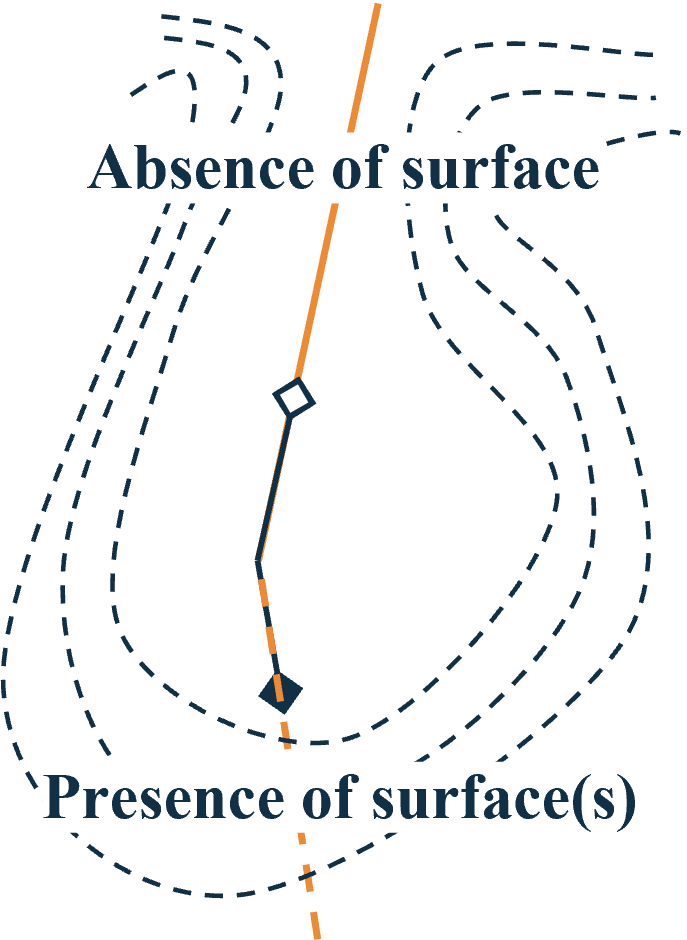}
    \caption{}
    \label{fig:transferloss_itutive}
  \end{subfigure}
    \caption{(a) Visibility as a continuous function of both point position $\bm{X}$ and sample direction $\bm{\omega}$ (top) leads to multiple queries and point-wise feature $F$ duplication. In contrast, by fixing the sampling directions $\bm{\omega}^{(n)}$ (bottom), it only requires a single query. We extensively analyze our visibility field simplification strategy over $10\%$ of training data and report average accuracy over the sample size $n$ in (b) with the following procedure: For each sample size decision, we sample 10k query points per model (same sampling distribution as training data) with visibility evaluated over $\bm{\omega}^{(n)}$, that we apply cosine distance interpolation (c) to infer 10k randomly sampled validation directions per point. For aggregating multi-view features (d), we apply the same interpolation method as in (c) for each camera view direction to evaluate visibility along the back projection ray. (e) Directional visibility naturally constraints on the presence of surface.}
    \label{}
\end{figure*}

\subsection{Visibility Field}
\label{sec:vis_field_definition}



For any point $\bm{X} \in \mathbb{R}^3$, whether it is visible $V \in \{0, 1\}$ along any view direction $\bm{\omega} \in \mathbb{R}^3$ of the unit sphere $S$ forms a continuous field and can be parameterized using MLP:
\begin{equation}
  \text{MLP}_{\phi} : (\bm{X}, \bm{\omega}) \to V_{\phi} \in [0, 1], \ \bm{\omega} \in S
  \label{eq:implicit_visibility_field}
\end{equation}
where $\phi$ is the MLP weights. Although \cref{eq:implicit_visibility_field} helps mitigate the cost of the ray integral, querying visibility over $s$ directions still requires $s$ calls per point of interest. This cost is exacerbated in pixel-aligned settings due to excessive point-wise feature $\bm{F}$ duplication as illustrated in \cref{fig:visibility_query} (top), leading to substantially high memory overhead.


We observe that, by uniformly sampling a discrete set of directions, visibility along any direction can be interpolated by \cref{eq:cosine_distance_interpolation} (\cref{fig:interpolation_fig}), with accuracy capped by the sample size. Thus, the $O(s)$ complexity can be reduced to a single MLP call plus an additional interpolation cost. To this end, we propose an alternative definition, by treating visibility as a function of $\bm{X}$ conditioned on a fixed set of $n$ sample directions $\bm{\omega}^{(n)}$ as in \cref{fig:visibility_query} (bottom).
\begin{equation}
  \text{MLP}_{\phi | \bm{\omega}^{(n)}} : (\bm{X}, \bm{F}) \to V_{\phi | \bm{\omega}^{(n)}} \in [0, 1]^n, \bm{\omega}^{(n)} \in S
  \label{eq:semi_implicit_visibility_field}
\end{equation}
In practice, we interpolate by top $k$ closest cosine distance with respect to $\bm{\omega}^{(n)}$:

\begin{equation}
\begin{aligned}
& \{\bm{\omega}^{(i)}\} _{i=1 \dots k}  = topk(\bm{\omega}^{(n)} \cdot \bm{\omega})\\
& V(\bm{X}, \bm{\omega}) = \sum_{i=1}^{k} \frac{V_{ \bm{\omega}^{(i)}}(\bm{X}) (\bm{\omega}^{(i)} \cdot \bm{\omega})}{\sum_{j=1}^k (\bm{\omega}^{(j)} \cdot \bm{\omega})}
\end{aligned}
  \label{eq:cosine_distance_interpolation}
\end{equation}

The simplification in \cref{eq:semi_implicit_visibility_field} leads to a surprisingly simple implementation--visibility prediction can be treated as $n-D$ binary classification supervised by BCE loss:
\begin{equation}
L_{\text{Vis}} = V_{\text{GT}} \cdot \log V_{\phi | \bm{\omega}^{(n)}} + (1 - V_{\text{GT}}) \cdot \log(1- V_{\phi | \bm{\omega}^{(n)}})
    \label{eq:bce_loss}
\end{equation}
%






\subsection{Framework Overview}
As shown in \cref{fig:framework}, our framework represents all fields using MLPs. Given sparse multi-view RGB-D frames $\{ (\mathcal{I}_i, \mathcal{D}_i), i=1 \dots m \}$ with known camera projection matrices $\pi_i$, we first extract depth point cloud voxelized 3D feature and pixel-aligned RGB 2D feature to directly predict visibility. Then we utilize inferred visibility to guide the aggregation of multi-view 2D feature, which are paired with 3D feature for joint geometry and albedo inference.

\begin{figure}
    \centering
    \includegraphics[width=0.9\linewidth]{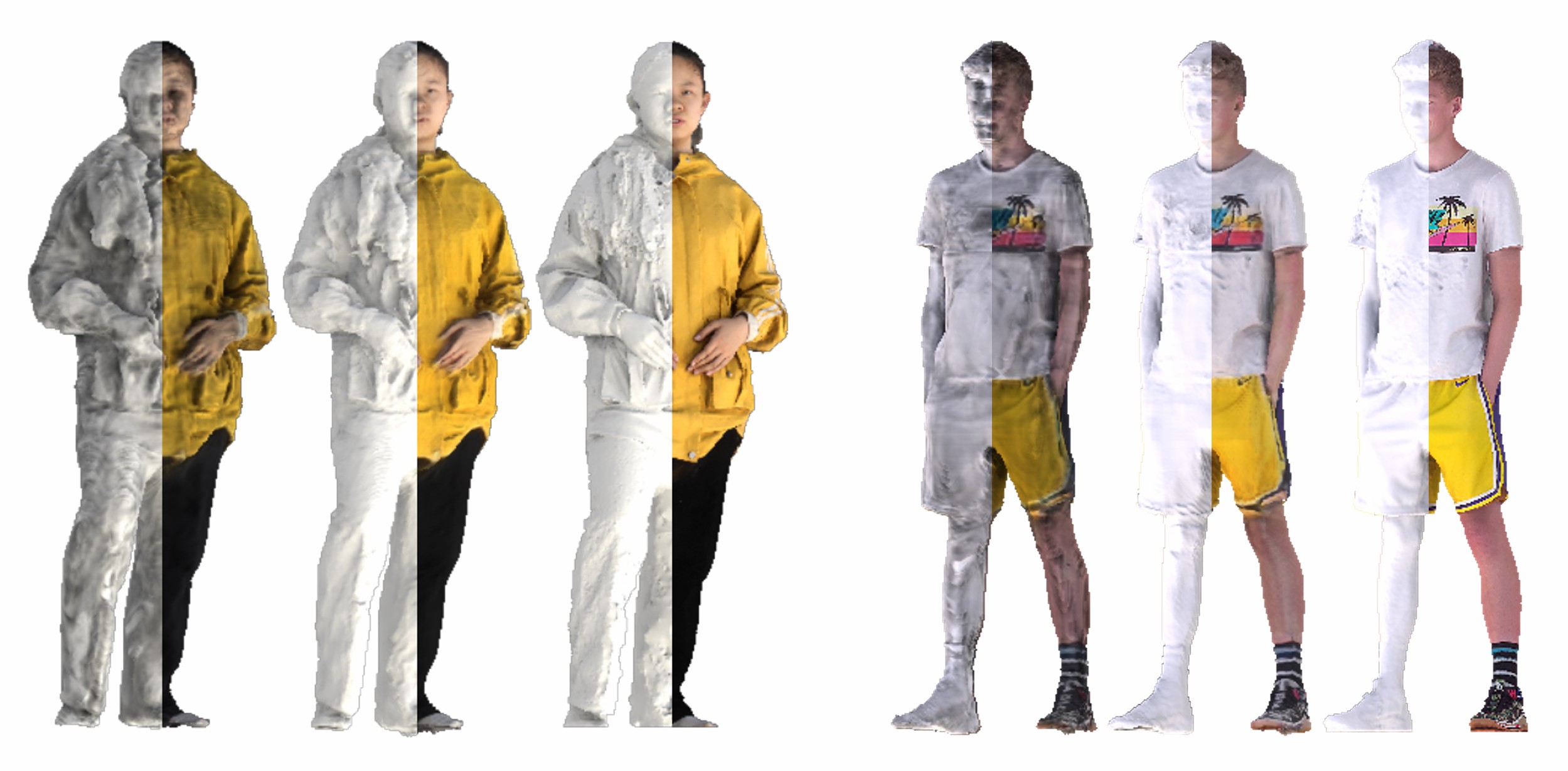}
    \vspace{-8pt}
    \caption{TransferLoss ablation: From left to right: w/o, w TransferLoss, ground truth. TransferLoss significantly mitigates rendering defects and improves near-surface visibility prediction accuracy ($27.5 \% \to 88.4 \%$ for synthesized testset, $20.8 \% \to 84.7 \%$ for real-captured data).}
    \label{exp:ablation_transfer_loss}
    \vspace{-0.5cm}
\end{figure}

\subsection{Hybrid Feature Extraction}
Conventionally, evaluating visibility requires tracing ray along the direction of interest and checking surface hits. In other words, it requires reasoning about the geometric feature of surfaces near that direction. Since we model both geometry (occupancy) and visibility, it is crucial to also bridge the two fields to incorporate their interconnection.

To this end, we adopt a hybrid feature extraction procedure, by separating depth from RGB and encoding it as 3D feature. Specifically, we follow ~\cite{peng2020convolutional} to unproject depth image into pointcloud, voxelize and filter using 3D convolutional as coarse feature volume. For a query point $\bm{X}$, we acquire its local 3D feature $F_{\mathcal{D}}$ by trilinear interpolation based on point coordinates and then share it with both visibility and occupancy MLPs. Intuitively, the 3D feature originates from voxelized point cloud, which can be viewed as noisy samples of the underlying surface (decision boundary of the occupancy field). By applying 3D convolution, the network reasons the coarse geometric surface feature over a sufficiently large receptive field, consequently aiding visibility inference. 3D feature is necessary for accurate visibility prediction as ablated in \cref{fig:3d_fea_visibility}.

For multi-view RGB frames, we directly filter them using HRNet \cite{sun2019deep} to acquire 2D feature maps $F_i$. The local 2D features are then extracted in the pixel-aligned fashion as in \cite{saito2019pifu}, by projecting the point coordinate $\bm{X}$ onto each view as the image coordinate $\pi(\bm{X})$, then bilinearly interpolating the corresponding feature maps $F_i(\pi(\bm{X}))$. Compared to feature volume, 2D feature maps have much higher resolution and thus grant better details.

Follow \cref{eq:semi_implicit_visibility_field}, we infer $V_{\phi | \bm{\omega}^{(n)}}$ by providing 3D feature $F_{\mathcal{D}}$ and averaged 2D feature $F_{avg}$:
\begin{equation}
  \text{MLP}_{\phi | \bm{\omega}^{(n)}} : (\bm{X}, F_{\mathcal{D}}, F_{avg}) \in [0, 1]^n
  \label{eq:visibility_inference}
\end{equation}

\begin{figure*}
    \centering
    \includegraphics[width=\linewidth]{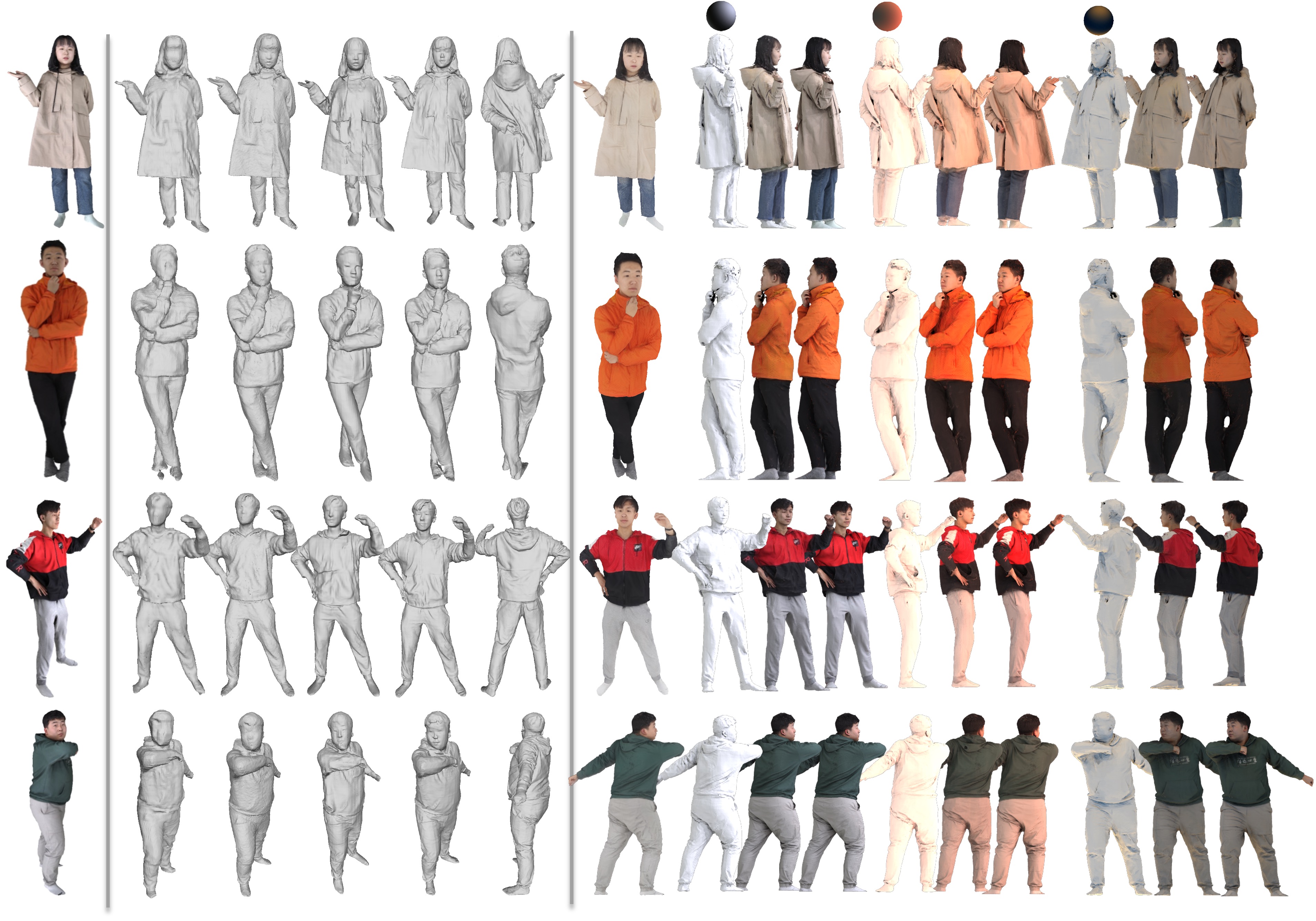}
    \vspace{-0.7cm}
    \begin{center}
        \footnotesize 
        \leftline{~\qquad(a)~~\quad\qquad(b)~~~~~\quad\quad(c)~\quad\qquad(d)~~~~~~\qquad\qquad(e)~~~~~\qquad\quad\qquad(f)~~~\qquad(g)~~~`~\qquad\qquad\qquad\quad(h)~~~~~~~\quad\qquad\qquad\qquad(i)}
    \end{center}  
    \vspace{-0.5cm}
    \caption{Qualitative comparisons on THUman2.0~\cite{yu2021function4d}. We show (a) shaded color reference, geometry of (b) MV-PIFu~\cite{saito2019pifu}, (c) Function4D~\cite{yu2021function4d},
    (d) MV-PIFuHD~\cite{saito2020pifuhd}, (e) ours, (f) our albedo, (g-i) from left to right: our irradiance, our relighting and ground truth.}
    \label{fig:thuman}
\end{figure*}

\subsection{Field Alignment Regularization}
\label{sec:transfer_loss}
Given similar formulations, it is natural to train the visibility field and occupancy field together. However, as the point moves across the surface (occupancy classification boundary), its visibility changes drastically, ranging from fully occluded to partially visible. A slight misalignment between the two fields could cause inner visibility leakage and introduce conspicuous rendering defects as in \cref{exp:ablation_transfer_loss}. To enforce their alignment, a common practice is to explicitly constrain their correspondence, by matching visibility with surface queried along the ray\cite{srinivasan2021nerv}, but at the expense of substantially large training overhead.

Since we bridge the two fields with the 3D feature reasoning about their correlation, we exploit the constraint of visibility over occupancy (\cref{fig:transferloss_itutive}) by emphasizing the accuracy of near-surface visibility prediction, so that their alignment can be implicitly enforced. To this end, we propose a novel TransferLoss inspired by radiance transfer in \cref{eq:transfer_function}:
\begin{equation}
    L_{\text{Transfer}} = \sum_{i} |\hat{T}(\bm{X}, V_{\phi | \bm{\omega}^{(i)}}) - T(\bm{X}, V_{\text{GT}})|
    \label{eq:transfer_loss}
\end{equation}

It supervises visibility, but (1) prioritizes the normal-facing directions in contrast to equal weighting in BCE (\cref{eq:bce_loss}) loss, and (2) follows the same parameterization for diffuse BRDF evaluation, which intuitively renders the light-independent part of the scene and perceptually penalizes the difference.

Since occupancy and visibility are jointly supervised, we directly sample query points on the ground truth mesh surface and use their normals to assist \cref{eq:transfer_loss} evaluation for inferred visibility. TransferLoss effectively enforces fields alignment, resulting in significantly less rendering defects and higher relighting fidelity as ablated in Sec.~\ref{exp:ablation} and \cref{exp:ablation_transfer_loss}.

\subsection{Visibility-Guided Feature Aggregation}
\label{sec:vis_guided_feature_aggregation}
For each view, we trace back to its camera center and evaluate the directional visibility as described in \cref{eq:cosine_distance_interpolation}. We then prioritize visible features over occluded ones using clamped negative log weighted average:
\begin{equation}
 \begin{aligned}
    &F_{agg} = \sum_{i=1}^m W_{V_{\phi | \bm{\omega}^{(n)}})} F_i(\pi_i(\bm{X})) \\
    &W_{V_{\phi | \bm{\omega}^{(n)}}} (\bm{X}, \bm{\omega}) = \max(-\log(1 - V_{\phi | \bm{\omega}^{(n)}}), 100)
    \label{eq:vis_guided_aggregation}
 \end{aligned}
\end{equation}


Thus, occupancy and albedo are represented as:
\begin{equation}
  \begin{aligned}
  \text{MLP}_{occ} : (\bm{X}, F_{\mathcal{D}}, F_{agg}) \to [0, 1] \\
  \text{MLP}_{albedo} : (\bm{X}, F_{\mathcal{D}}, F_{agg}) \to \mathbb{R}^3
  \end{aligned}
  \label{eq:occupancy_albedo}
\end{equation}

\subsection{Reconstruction and Relighting}
During inference, we first apply visibility-guided aggregation to sample a grid of occupancy for surface mesh extraction \cite{lorensen1987marching}. We then pass the mesh vertices to obtain albedo and visibility using the same technique. Following \cref{eq:transfer_function}, we compute per-vertex transfer coefficients and directly apply them for rasterized self-shadowed relighting.




\begin{figure*}
    \centering
    \includegraphics[width=\linewidth]{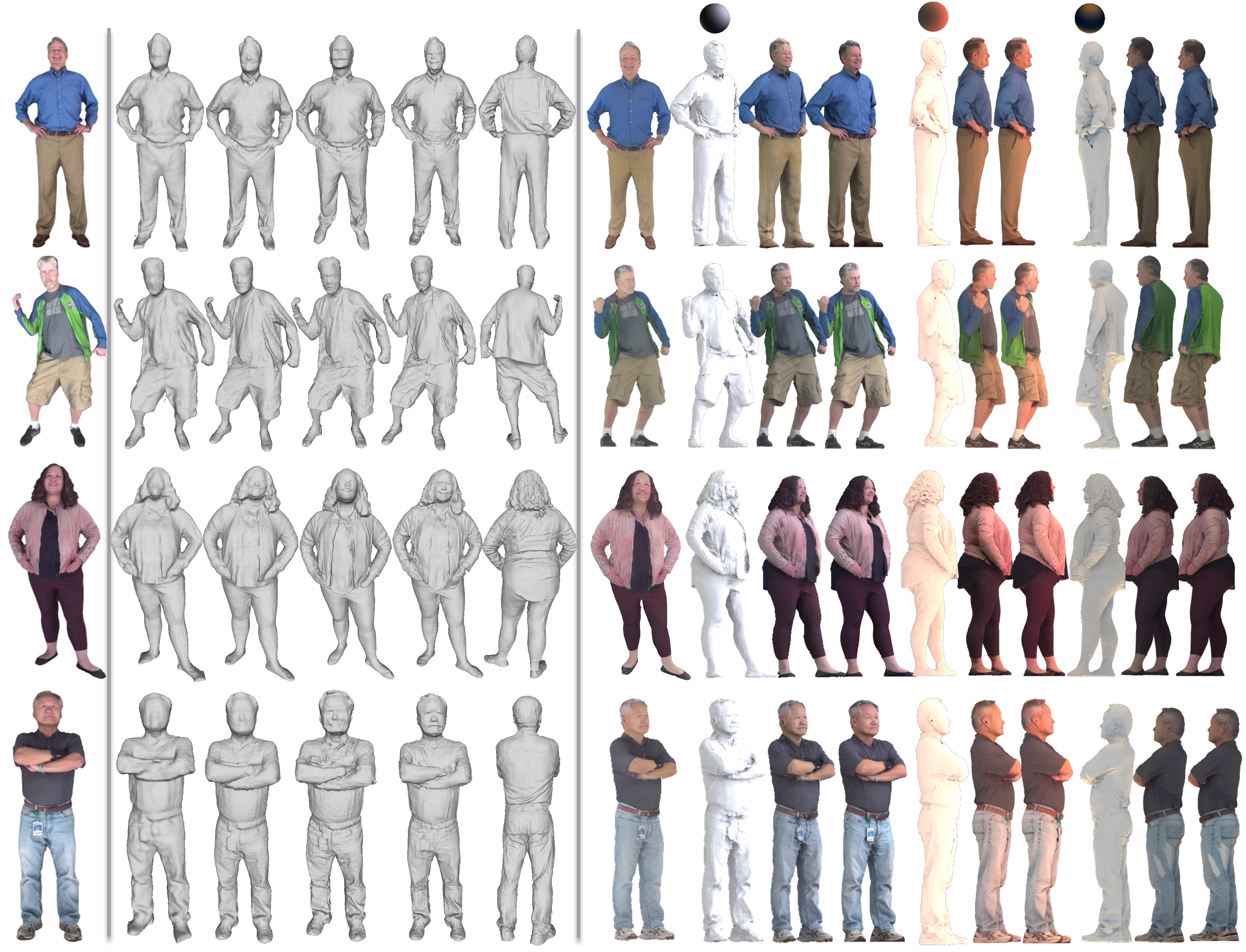}
    \vspace{-0.7cm}
    \begin{center}
        \footnotesize 
          \begin{center}
        \footnotesize 
        \leftline{~\qquad(a)~~~~\quad\qquad(b)~~~~~\quad\quad(c)~\quad\qquad(d)~~~~~~~~~\qquad\qquad(e)~~~~~~~~\qquad\quad\qquad(f)~~~~~~\qquad(g)~~~~~~~\qquad\qquad\qquad\quad(h)~~~~~~~\qquad\qquad\qquad(i)}
    \end{center}  
    \vspace{-0.5cm}
    \caption{Qualitative comparisons on Twindom. From left to right: (a) shaded color reference, geometry of (b) MV-PIFu~\cite{saito2019pifu}, (c) Function4D~\cite{yu2021function4d},
    (d) MV-PIFuHD~\cite{saito2020pifuhd}, (e) ours, (f) our albedo, (g-i) from left to right: our irradiance, our relighting and ground truth.}
    \end{center}
    \label{fig:twindom}
\end{figure*}






\begin{figure}
  \centering
  \begin{subfigure}{0.47\linewidth}
    \includegraphics[width=\linewidth]{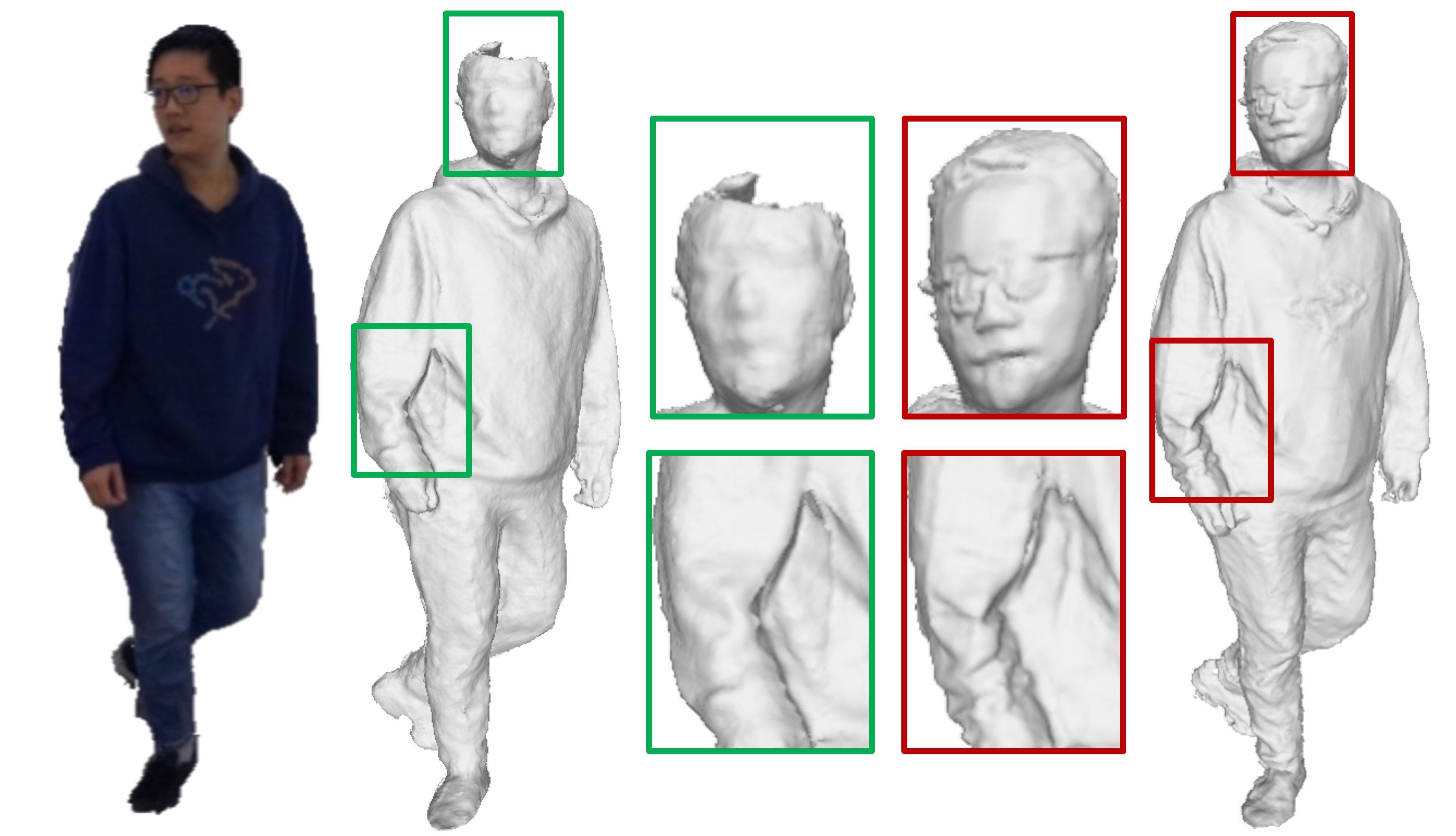}
    \label{fig:kinect_case1}
  \end{subfigure}
  \begin{subfigure}{0.47\linewidth}
    \includegraphics[width=\linewidth]{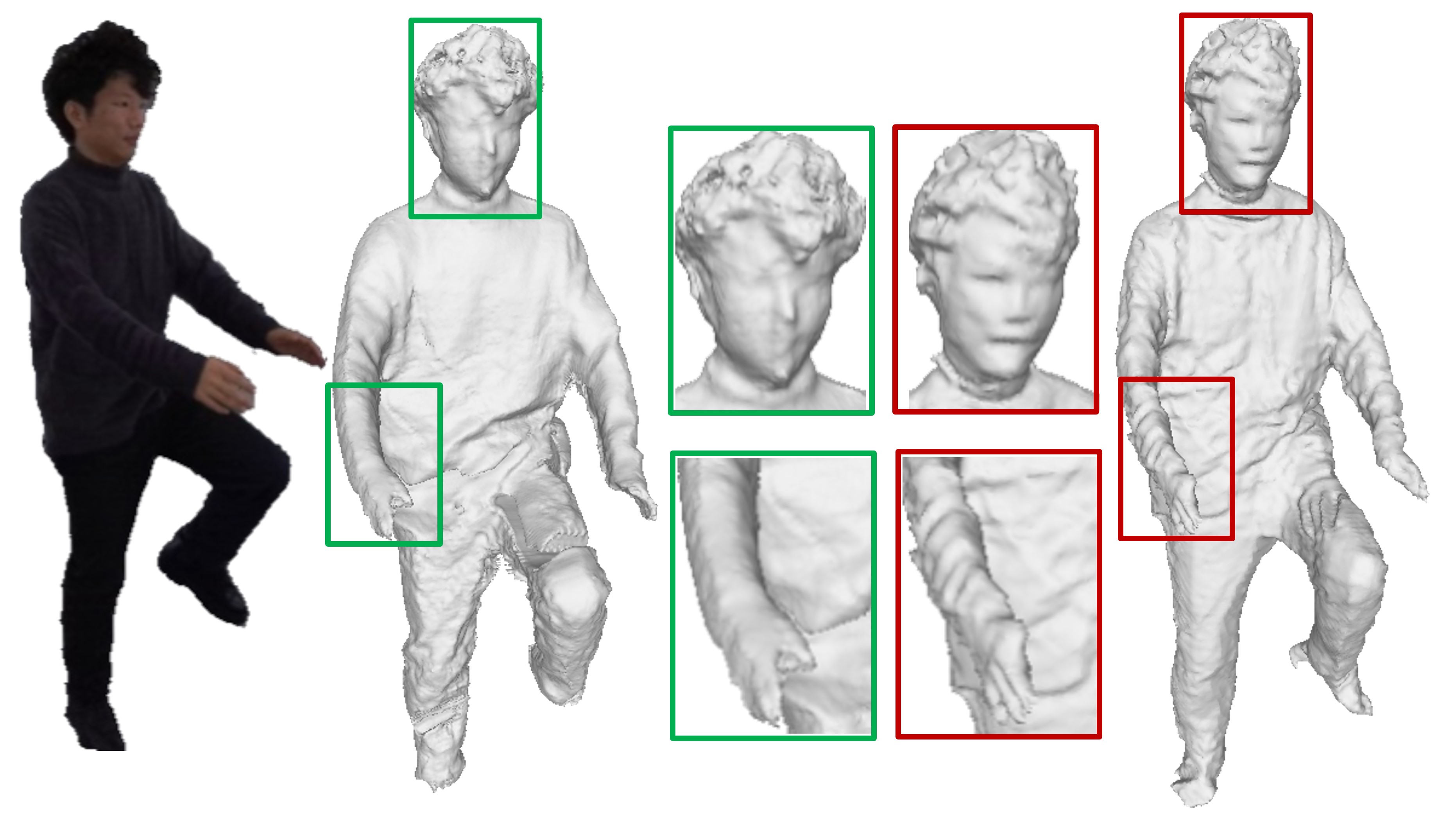}
    \label{fig:kinect_case2}
  \end{subfigure}
  \vspace{-0.1cm}
  \caption{Visualization comparison between Function4D~\cite{yu2021function4d} (green) and our method (red) on data captured by Kinect.
  }
  \label{exp:kinect}
  \vspace{-0.5cm}
\end{figure}

\section{Experiments}
\label{sec:exp}
\subsection{Experimental Settings}
\label{exp:setting}

\noindent\textbf{Training details.} For training, we collect 400 high-quality clothed human scans from THUman2.0~\cite{yu2021function4d}, rotate each one around the yaw axis, and apply random shifts to obtain 60 views. For each view, we render 512$\times$512 images of albedo, color using diffuse BRDF and depth fused with TOF depth sensors noise \cite{fankhauser2015kinect}. To simulate the incomplete depth caused by capture insensitivity for materials such as hair, we use \cite{hairSeg} to mask hair out.

For each scanned mesh, we sample total $5$k points for occupancy and visibility, with $4$k near surface and $1$k uniformly within the bounding volume. Near-surface points are sampled using normal distribution with standard deviation of $0.05$, and we ensure that half of them with distances less than the standard deviation are used for albedo training. As described in~\cref{sec:vis_field_definition} and \cref{fig:sample_curve}, we uniformly sample 64 fixed directions using the Fibonacci lattice and keep it consistent throughout the experiment. We use Embree \cite{wald2014embree} for ground truth visibility evaluation.

In addition to visibility, we supervise per-sample occupancy with BCE loss and albedo with $L_1$ loss. We also extract patches using depth-guided raymarching \cite{lin2022efficient} and supervise its albedo using VGG perceptual loss.

We set the view number to 4 and train our framework using Adam \cite{kingma2014adam} optimizer and Cyclic learning rate (lr) scheduler \cite{smith2017cyclical} over 600 epochs. The lr ranges from $5e-5$ to $5e-4$ every 5 epochs, with the max lr halved every 100 epochs. 

\noindent\textbf{Evaluation details.}
For evaluation, we prepare 100 training-excluded scans from THUman2.0, and an additional 100 scans from Twindom \cite{shao2022doublefield}. We follow the same rendering procedure as for training and report metrics averaged over the two sets. We further prepare real-captured RGB-D video sequences from a synchronized multi-Kinect capturing system that we leverage RVM~\cite{lin2022RVM} for foreground mask segmentation. For all experiments, we keep the number of views to 4 except for the view-number ablation. All meshes are extracted from ${512}^3$ voxel using Marching Cube \cite{lorensen1987marching}. The relighting results are rendered using diffuse BRDF with inferred visibility and SH order of 2.

All experiments run on a PC with an Nvidia GeForce RTX3090 GPU and an Intel i7-8700k CPU. 
Our framework requires $50$ ms for 3D and $100$ ms for 2D feature extraction, and $200$ ms per 2 million point queries for each of the 4-layer ResNet MLP decoders. The total reconstruction process takes approximately 3 seconds.
After experimenting with TensorRT conversion, 3D and 2D feature extraction can be reduced to $10$ and $7$ ms, respectively. Since they can be performed in parallel, our framework has the potential to achieve real-time performance under heavy decoder optimization.

\begin{table}[t]\footnotesize
    \centering
    \resizebox{0.48\textwidth}{!}{
    \begin{tabular}{l|lll}
    \toprule[0.8pt]
        \multicolumn{1}{c|}{\multirow{2}{*}{Methods}} &
        \multicolumn{3}{c}{Metrics} \\
            \makecell[c]{} & \makecell[c]{NC $\uparrow$} & \makecell[c]{CD ($L_1$) $\downarrow$} & \makecell[c]{F-score (0.5\%)$\uparrow$} \\
        \midrule[0.3pt]
            \makecell[c]{MV-PIFu~\cite{saito2019pifu}} & 
            \makecell[c]{0.912}  & 
            \makecell[c]{0.145} & 
            \makecell[c]{0.624} \\
            \makecell[c]{MV-PIFuHD~\cite{saito2020pifuhd}} & 
            \makecell[c]{0.906} &
            \makecell[c]{0.135} & 
            \makecell[c]{0.690} \\
            \makecell[c]{Function4D~\cite{yu2021function4d}} & 
            \makecell[c]{0.893} &  
            \makecell[c]{0.129} &  
            \makecell[c]{0.704} \\
            \makecell[c]{Ours} & 
            \makecell[c]{\textbf{0.917}} &  
            \makecell[c]{\textbf{0.122}} & 
            \makecell[c]{\textbf{0.736}} \\
        \bottomrule[0.8pt]
    \end{tabular}}
    \caption{Quantitative comparisons on reconstruction quality.}
    \label{tab:geo_comparison}
\end{table}

\noindent\textbf{Metrics.} We report normal consistency (NC) \cite{mescheder2019occupancy}, $L_1$ Chamfer Distance (CD $L_1$) \cite{fan2017point} and F-score \cite{knapitsch2017tanks} for geometric quality evaluation.
Specifically, NC is calculated as the mean absolute dot product of normals of points sampled from reconstructed mesh and their closest neighbours' ones on ground truth mesh.
We follow \cite{mescheder2019occupancy} to use 1/10 of the maximum bounding box edge as unit 1 for CD $L_1$ and 0.5\% as the F-score distance threshold as suggested by \cite{tatarchenko2019single}.

To evaluate the quality of relighting, we adopt the peak signal-to-noise ratio (PSNR), the structural similarity index measure (SSIM) \cite{wang2004image} and the learned perceptual image patch similarity (LPIPS) \cite{zhang2018unreasonable}. We mask out renderings with ground truth alpha channel and only report the average contributions of valid pixels.

\noindent\subsection{Comparisons}
\noindent\textbf{Reconstruction.} We compare our framework with the state-of-the-art prior-free sparse-view reconstruction approaches, namely multi-view PIFu (MV-PIFu) \cite{saito2019pifu}, multi-view PIFuHD (MV-PIFuHD) \cite{saito2020pifuhd} and Function4D \cite{yu2021function4d}. MV-PIFu takes RGB inputs and aggregates using averaging, which we adapt with additional depth input to ensure comparison consistency. PIFuHD integrates normal information, coarse-to-fine two-stage inference, and higher $1024\times1024$ resolution input. We self-implement multi-view RGB-D variants (MV-PIFuHD) with the same averaging aggregation as MV-PIFu. Normal maps are still inferred rather than being computed from depth maps due to noise concern. Function4D uses averaging in geometry inference as well, but its integration of the truncated PSDF serves as a strong signal to identify visible features and has been shown to generalize well on real captured data. For a fair comparison, we re-render and train all three approaches on our dataset until converge. Regrettably, we could only compare the geometry, since none of them estimates surface albedo.

\noindent\textbf{Relighting.} Limited by the dataset, which maps human into single albedo texture without differentiating specular components and coefficients, we were only able to render using diffuse BRDF. This also limits our relighting comparison with NeRF-like method \cite{boss2021nerd, chen2022relighting4d, srinivasan2021nerv}, where the view-dependent specular term is explicitly modeled and is crucial for rendering fidelity. Therefore, we directly compare our results to ray-traced ground truth to demonstrate our relighting performance.
\begin{figure}
  \centering
  \begin{subfigure}{0.45\linewidth}
    \centering
    \includegraphics[width=\linewidth]{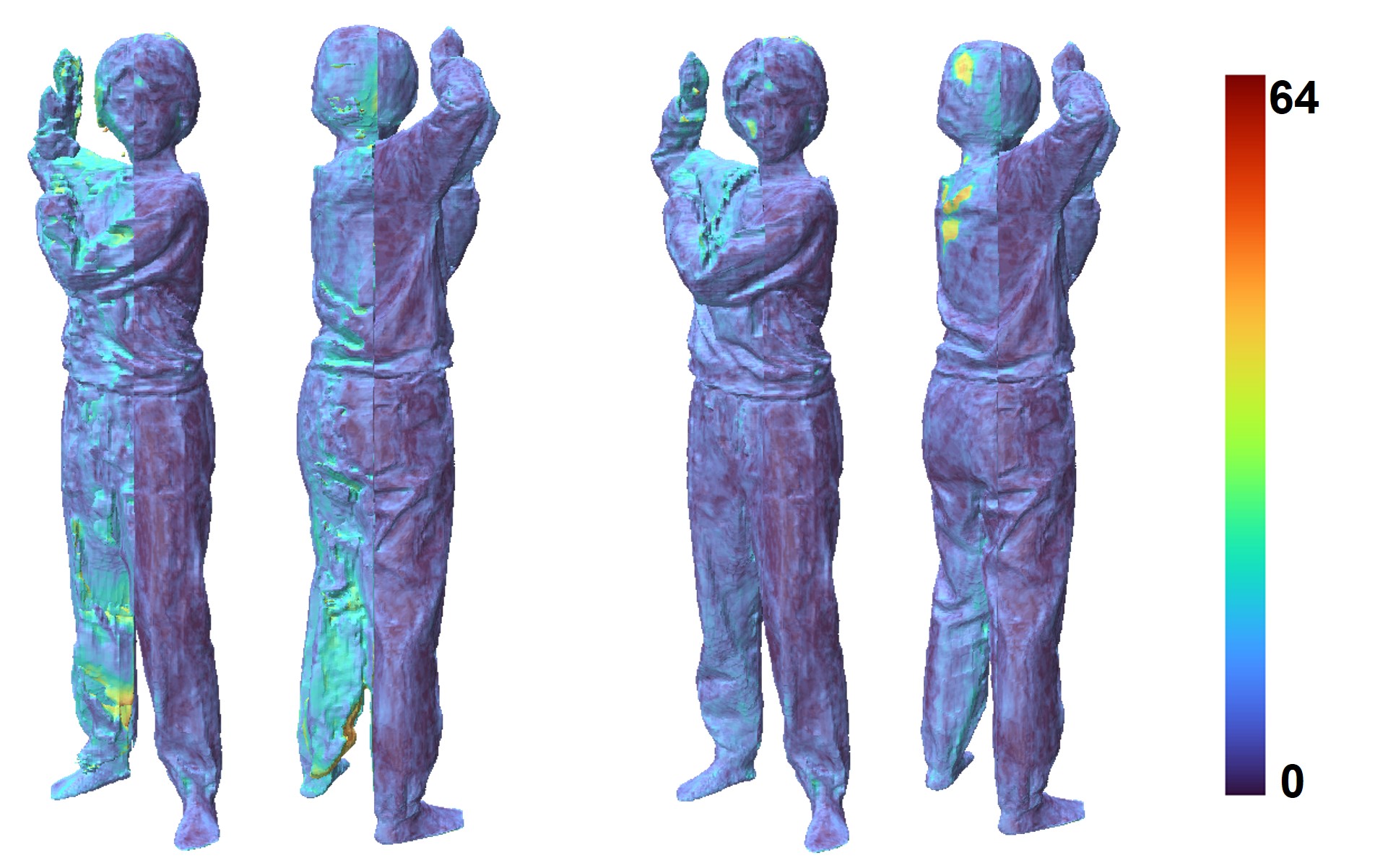} 
    \caption{}
    \label{fig:vis_error}
  \end{subfigure}
  \begin{subfigure}{0.45\linewidth}
  \centering
    \includegraphics[width=\linewidth]{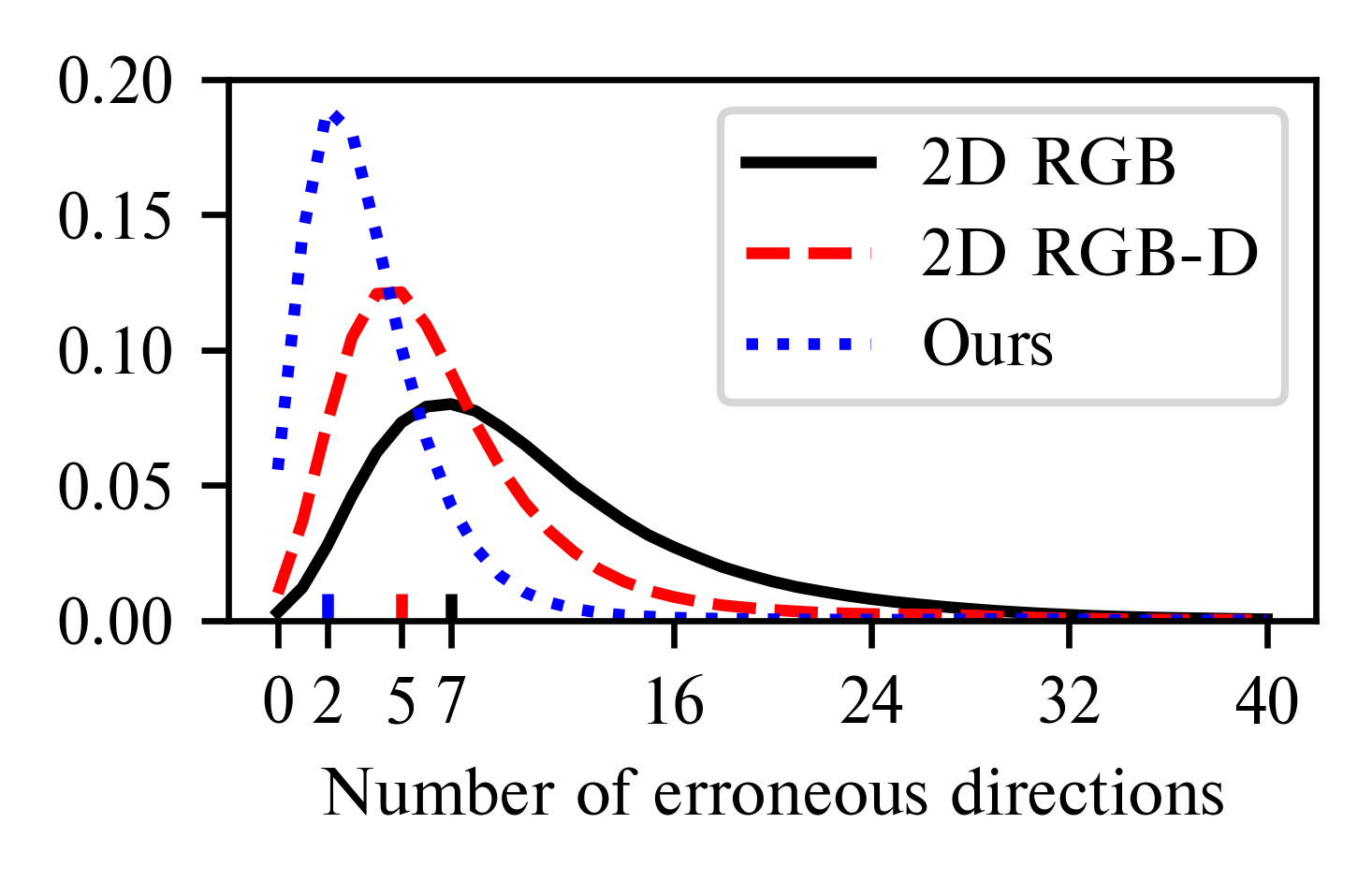}
    \caption{}
    \label{fig:vis_distribution}
  \end{subfigure}
  \caption{3D Feature ablation. (a) We compare the visibility error maps using 2D RGB input (left half of first two), 2D RGB-D input (left half of last two) and ours (right half of all four). (b) We count per-vertex number of wrong directional visibility predictions (out of 64) and plot the normalized histogram.}
  \label{fig:3d_fea_visibility}
\end{figure}

\noindent\textbf{Qualitative Comparison.} \cref{fig:thuman} shows the reconstruction on the synthetic datasets. Benefiting from visibility-guided feature aggregation, our method produces at least comparable details to MV-PIFuHD, but with simpler architecture, less input information and visual cues. Compared to over-smoothed geometry from Function4D, our hybrid features relax the dominant contribution of noisy depth input, leading to improved facial details.
We further demonstrate our generalizability on real captured data in comparison with Function4D. As shown in \cref{exp:kinect}, our method evidently produces more complete and detailed reconstructions, especially in regions of the eyes, glasses and hair.

\begin{table}[t]\footnotesize
\centering
\caption{Feature aggregation ablation.}
\label{tab:agg_ablation}
\resizebox{0.48\textwidth}{!}{
\begin{tabular}{l|lll|lll}
\toprule[0.8pt]
    \multicolumn{1}{c|}{\multirow{2}{*}{Method}} & 
    \multicolumn{3}{c|}{Geometry Metric} &
    \multicolumn{3}{c}{Relit Rendering} \\
    \makecell[c]{} &
    \makecell[c]{NC$\uparrow$} & 
    \makecell[c]{CD ($L_1$)$\downarrow$} &
    \makecell[c]{F-score (0.5\%)$\uparrow$} &
    \makecell[c]{PSNR$\uparrow$} &
    \makecell[c]{SSIM$\uparrow$} &
    \makecell[c]{LPIPS$\downarrow$} \\
\midrule[0.3pt]
    \makecell[c]{Average} &
    \makecell[c]{0.911}  & 
    \makecell[c]{0.129} & 
    \makecell[c]{0.696}  & 
    \makecell[c]{17.933} & 
    \makecell[c]{0.650}  & 
    \makecell[c]{0.327}\\
    \makecell[c]{Attention} &
    \makecell[c]{0.909}  & 
    \makecell[c]{0.127} & 
    \makecell[c]{0.710}  & 
    \makecell[c]{19.148} & 
    \makecell[c]{0.713}  & 
    \makecell[c]{0.259}\\
    \makecell[c]{Ours} & 
    \makecell[c]{\textbf{0.917}} &  
    \makecell[c]{\textbf{0.122}} & 
    \makecell[c]{\textbf{0.736}} &
    \makecell[c]{\textbf{23.436}} & 
    \makecell[c]{\textbf{0.809}}  & 
    \makecell[c]{\textbf{0.196}}\\
\bottomrule[0.8pt]
\end{tabular}}
\end{table}

\noindent\textbf{Quantitative Comparison} summarized in \cref{tab:geo_comparison} is consistent with qualitative analysis, and our method outperforms others in all metrics.

\begin{figure}
    \centering
    \includegraphics[width=\linewidth]{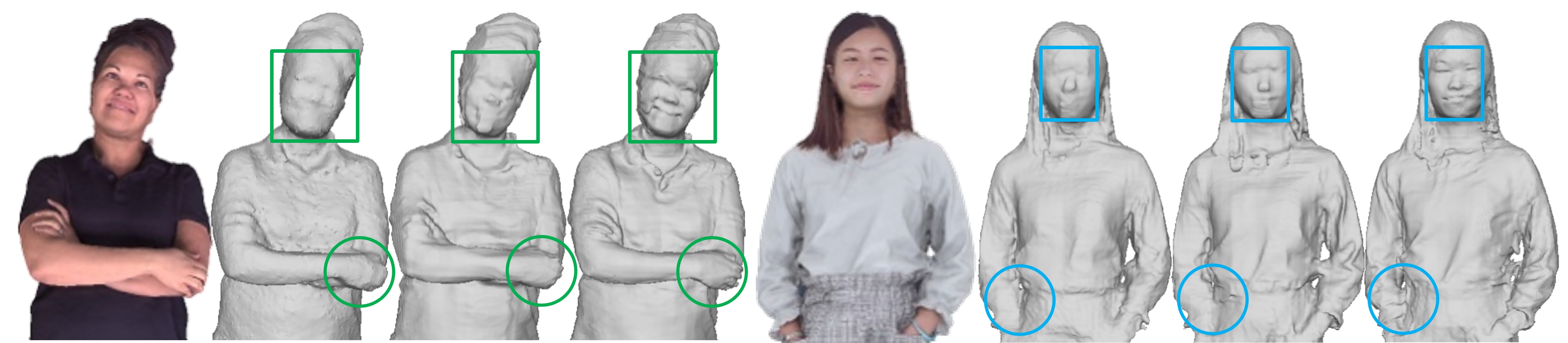}
    \caption{Feature aggregation ablation in geometry. From left to right: color reference, the results of average, attention and ours.}
    \label{exp:abla_aggreation_geo}
\end{figure}

\begin{figure}
  \centering
  \begin{subfigure}{0.45\linewidth}
    \includegraphics[width=\linewidth]{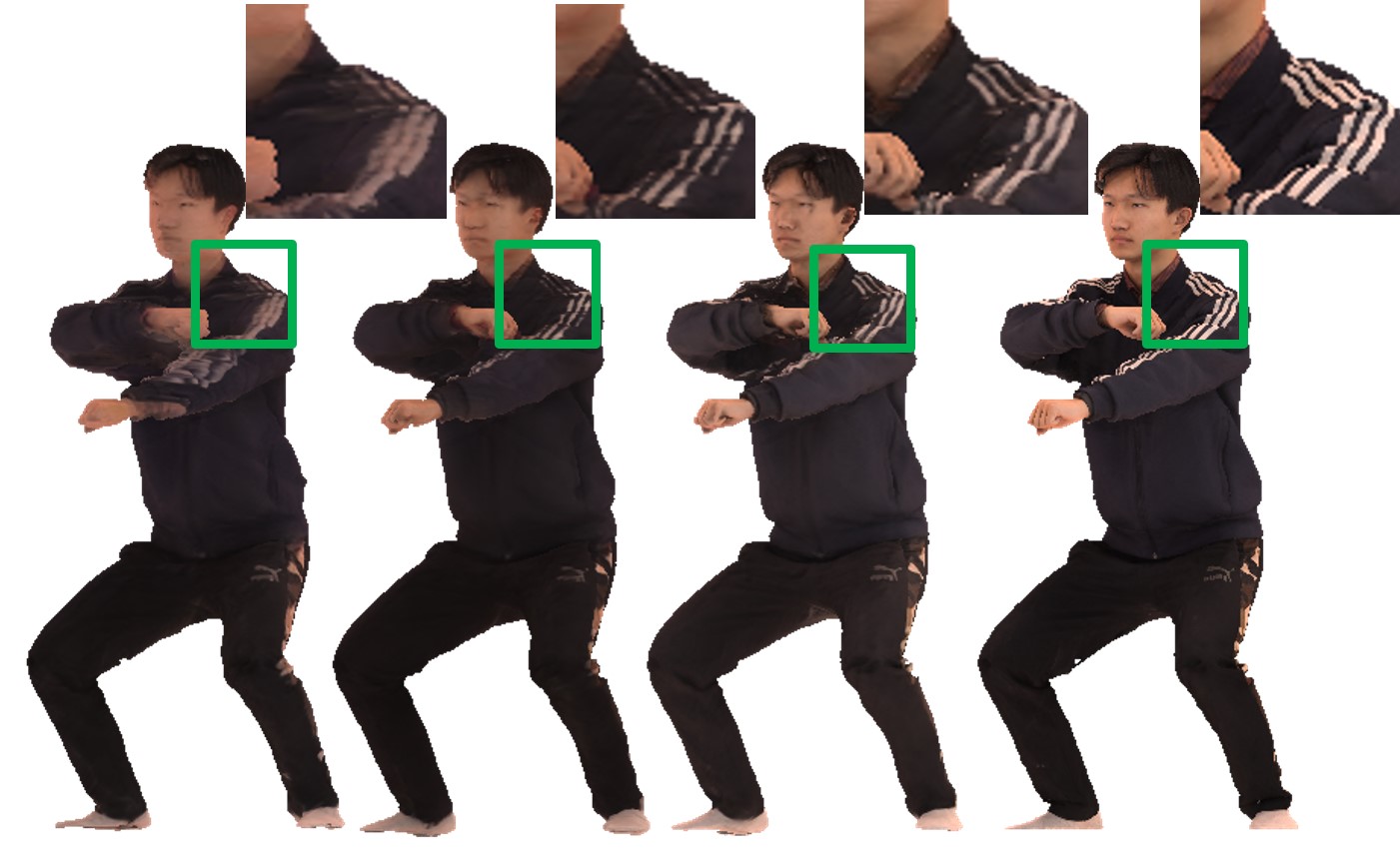}
    \label{fig:abla_render_a}
  \end{subfigure}
  \begin{subfigure}{0.40\linewidth}
    \includegraphics[width=0.98\linewidth]{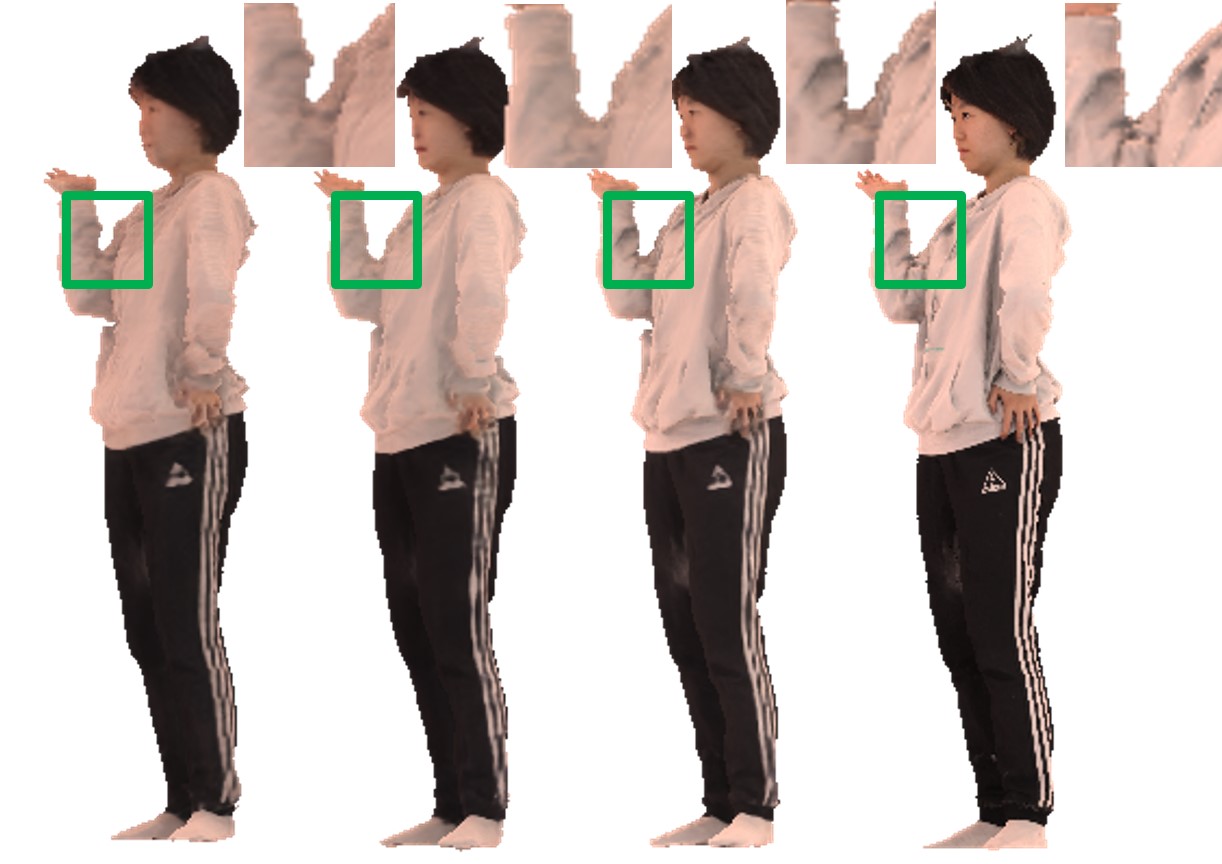}
      \label{fig:abla_render_b}
  \end{subfigure}
  \vspace{-20pt}
  \caption{Feature aggregation ablation in rendering quality. From left to right: average, attention, and ours and ground truth. 
  }
  \label{fig:abla_render}
\end{figure}

\subsection{Ablation Study}
\label{exp:ablation}
\noindent\textbf{3D Feature.} We ablate our hybrid feature with 2D RGB and 2D RGB-D variants. The results validate the necessity of the 3D feature, as it predicts visibility with higher accuracy (\cref{fig:vis_error}) and lower error variance (\cref{fig:vis_distribution}).

\noindent\textbf{Visibility-guided Feature Aggregation.} In comparison with other aggregation techniques, namely averaging \cite{saito2019pifu} and self-attention \cite{yu2021function4d}, we implement them in our framework. Our strategy surpasses others in metrics (\cref{tab:agg_ablation}) and achieves sharper geometric details (\cref{exp:abla_aggreation_geo}), better rendering fidelity (\cref{fig:abla_render}), even near heavily folded clothing thanks to our occlusion-aware aggregation.

\begin{figure}
    \centering
    \includegraphics[width=\linewidth]{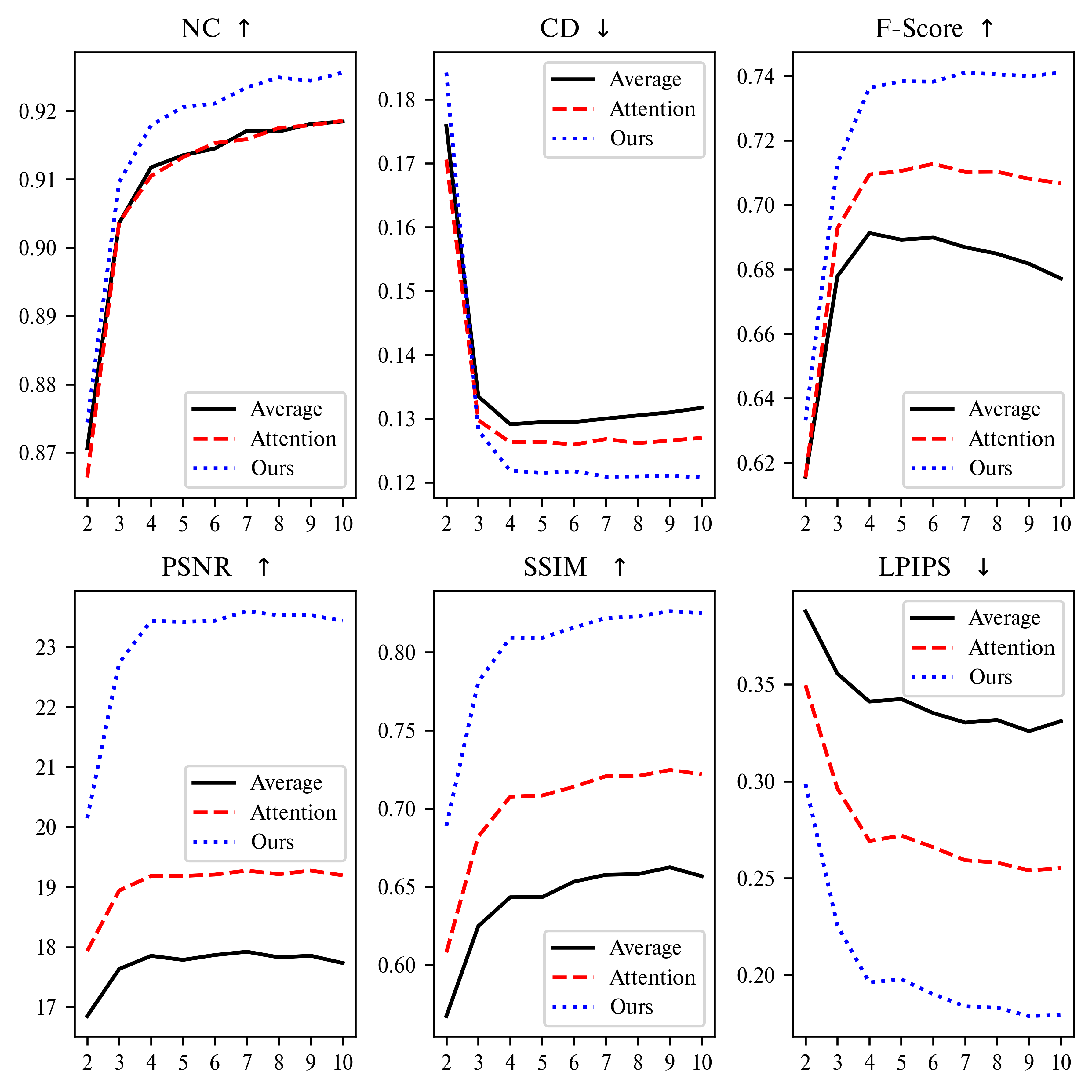}
    \vspace{-20pt} 
    \caption{View number ablation.}
    \label{exp:ablation_view_num}
\end{figure}

\noindent\textbf{View number.} Though we train our model with 4 views, it generalizes well across different view numbers and achieves similar performance under sufficient view coverage as ablated in \cref{exp:ablation_view_num}. As view number increases, compared to degraded reconstruction quality from average and attention, our aggregation technique mitigates visibility ambiguity to leverage additional visual cues, resulting in better geometry.

\noindent\textbf{TransferLoss.} Finally, we verify the effectiveness of our proposed TransferLoss in \cref{exp:ablation_transfer_loss}. It enforces the alignment of the fields and significantly alleviates rendering defects.

\section{Conclusion}
\label{sec:conclusion}
\noindent\textbf{Limitations.} Our method cannot achieve real-time performance for interactive applications. We leave its acceleration using TensorRT and CUDA for future work. Moreover, our framework relies on depth input for accurate visibility prediction, it would be interesting to see if it can be extended to simpler RGB setup.

\noindent\textbf{Conclusion.}
In this work, we integrate visibility into sparse-view reconstruction framework by exploiting its effective guidance in multi-view feature aggregation and direct support for self-shadowed relighting. Our discretization strategy and novel TransferLoss enable visibility to be learned jointly alongside occupancy in an end-to-end manner. Our paper demonstrates the effectiveness of visibility in simultaneous reconstruction and relighting and provides a good baseline for future work.

\noindent\textbf{Acknowledgement.} This work was supported by NSFC No.62171255, National Key R\&D Program of China No.2022YFF0902201, Guoqiang Institute of Tsinghua University No.2021GQG0001 and Tsinghua-Toyota Joint Research Fund.

{\small
\bibliographystyle{ieee_fullname}
\bibliography{egbib}

\begin{thebibliography}{10}\itemsep=-1pt

\bibitem{alldieck2022photorealistic}
Thiemo Alldieck, Mihai Zanfir, and Cristian Sminchisescu.
\newblock Photorealistic monocular 3d reconstruction of humans wearing
  clothing.
\newblock In {\em Proceedings of the IEEE/CVF Conference on Computer Vision and
  Pattern Recognition}, pages 1506--1515, 2022.

\bibitem{bhatnagar2020ipnet}
Bharat~Lal Bhatnagar, Cristian Sminchisescu, Christian Theobalt, and Gerard
  Pons-Moll.
\newblock Combining implicit function learning and parametric models for 3d
  human reconstruction.
\newblock In {\em European Conference on Computer Vision}, pages 311--329.
  Springer, 2020.

\bibitem{boss2021nerd}
Mark Boss, Raphael Braun, Varun Jampani, Jonathan~T Barron, Ce Liu, and Hendrik
  Lensch.
\newblock Nerd: Neural reflectance decomposition from image collections.
\newblock In {\em Proceedings of the IEEE/CVF International Conference on
  Computer Vision}, pages 12684--12694, 2021.

\bibitem{chen2022relighting4d}
Zhaoxi Chen and Ziwei Liu.
\newblock Relighting4d: Neural relightable human from videos.
\newblock In {\em European Conference on Computer Vision}, pages 606--623.
  Springer, 2022.

\bibitem{chen2019learning}
Zhiqin Chen and Hao Zhang.
\newblock Learning implicit fields for generative shape modeling.
\newblock In {\em Proceedings of the IEEE/CVF Conference on Computer Vision and
  Pattern Recognition}, pages 5939--5948, 2019.

\bibitem{chibane2020implicit}
Julian Chibane, Thiemo Alldieck, and Gerard Pons-Moll.
\newblock Implicit functions in feature space for 3d shape reconstruction and
  completion.
\newblock In {\em Proceedings of the IEEE/CVF Conference on Computer Vision and
  Pattern Recognition}, pages 6970--6981, 2020.

\bibitem{debevec2000pursuing}
Paul Debevec.
\newblock Pursuing reality with image-based modeling, rendering, and lighting.
\newblock In {\em European Workshop on 3D Structure from Multiple Images of
  Large-Scale Environments}, pages 1--16. Springer, 2000.

\bibitem{fan2017point}
Haoqiang Fan, Hao Su, and Leonidas~J Guibas.
\newblock A point set generation network for 3d object reconstruction from a
  single image.
\newblock In {\em Proceedings of the IEEE conference on computer vision and
  pattern recognition}, pages 605--613, 2017.

\bibitem{fanello2014learning}
Sean~Ryan Fanello, Cem Keskin, Shahram Izadi, Pushmeet Kohli, David Kim, David
  Sweeney, Antonio Criminisi, Jamie Shotton, Sing~Bing Kang, and Tim Paek.
\newblock Learning to be a depth camera for close-range human capture and
  interaction.
\newblock {\em ACM Transactions on Graphics (TOG)}, 33(4):1--11, 2014.

\bibitem{fankhauser2015kinect}
P{\'e}ter Fankhauser, Michael Bloesch, Diego Rodriguez, Ralf Kaestner, Marco
  Hutter, and Roland Siegwart.
\newblock Kinect v2 for mobile robot navigation: Evaluation and modeling.
\newblock In {\em 2015 International Conference on Advanced Robotics (ICAR)},
  pages 388--394. IEEE, 2015.

\bibitem{feng2022fof}
Qiao Feng, Yebin Liu, Yu-Kun Lai, Jingyu Yang, and Kun Li.
\newblock Fof: learning fourier occupancy field for monocular real-time human
  reconstruction.
\newblock {\em arXiv preprint arXiv:2206.02194}, 2022.

\bibitem{guo2019relightables}
Kaiwen Guo, Peter Lincoln, Philip Davidson, Jay Busch, Xueming Yu, Matt Whalen,
  Geoff Harvey, Sergio Orts-Escolano, Rohit Pandey, Jason Dourgarian, et~al.
\newblock The relightables: Volumetric performance capture of humans with
  realistic relighting.
\newblock {\em ACM Transactions on Graphics (ToG)}, 38(6):1--19, 2019.

\bibitem{guo2015robust}
Kaiwen Guo, Feng Xu, Yangang Wang, Yebin Liu, and Qionghai Dai.
\newblock Robust non-rigid motion tracking and surface reconstruction using l0
  regularization.
\newblock In {\em Proceedings of the IEEE International Conference on Computer
  Vision}, pages 3083--3091, 2015.

\bibitem{he2021arch++}
Tong He, Yuanlu Xu, Shunsuke Saito, Stefano Soatto, and Tony Tung.
\newblock Arch++: Animation-ready clothed human reconstruction revisited.
\newblock In {\em Proceedings of the IEEE/CVF International Conference on
  Computer Vision}, pages 11046--11056, 2021.

\bibitem{hong2021stereopifu}
Yang Hong, Juyong Zhang, Boyi Jiang, Yudong Guo, Ligang Liu, and Hujun Bao.
\newblock Stereopifu: Depth aware clothed human digitization via stereo vision.
\newblock In {\em Proceedings of the IEEE/CVF Conference on Computer Vision and
  Pattern Recognition}, pages 535--545, 2021.

\bibitem{huang2021dynamic}
Buzhen Huang, Yuan Shu, Tianshu Zhang, and Yangang Wang.
\newblock Dynamic multi-person mesh recovery from uncalibrated multi-view
  cameras.
\newblock In {\em 2021 International Conference on 3D Vision (3DV)}, pages
  710--720. IEEE, 2021.

\bibitem{huang2020arch}
Zeng Huang, Yuanlu Xu, Christoph Lassner, Hao Li, and Tony Tung.
\newblock Arch: Animatable reconstruction of clothed humans.
\newblock In {\em Proceedings of the IEEE/CVF Conference on Computer Vision and
  Pattern Recognition}, pages 3093--3102, 2020.

\bibitem{ji2022geometry}
Chaonan Ji, Tao Yu, Kaiwen Guo, Jingxin Liu, and Yebin Liu.
\newblock Geometry-aware single-image full-body human relighting.
\newblock In {\em Computer Vision--ECCV 2022: 17th European Conference, Tel
  Aviv, Israel, October 23--27, 2022, Proceedings, Part XVI}, pages 388--405.
  Springer, 2022.

\bibitem{kajiya1986rendering}
James~T Kajiya.
\newblock The rendering equation.
\newblock In {\em Proceedings of the 13th annual conference on Computer
  graphics and interactive techniques}, pages 143--150, 1986.

\bibitem{kanamori2019relighting}
Yoshihiro Kanamori and Yuki Endo.
\newblock Relighting humans: occlusion-aware inverse rendering for full-body
  human images.
\newblock {\em arXiv preprint arXiv:1908.02714}, 2019.

\bibitem{kingma2014adam}
Diederik~P Kingma and Jimmy Ba.
\newblock Adam: A method for stochastic optimization.
\newblock {\em arXiv preprint arXiv:1412.6980}, 2014.

\bibitem{knapitsch2017tanks}
Arno Knapitsch, Jaesik Park, Qian-Yi Zhou, and Vladlen Koltun.
\newblock Tanks and temples: Benchmarking large-scale scene reconstruction.
\newblock {\em ACM Transactions on Graphics (ToG)}, 36(4):1--13, 2017.

\bibitem{lagunas2021single}
Manuel Lagunas, Xin Sun, Jimei Yang, Ruben Villegas, Jianming Zhang, Zhixin
  Shu, Belen Masia, and Diego Gutierrez.
\newblock Single-image full-body human relighting.
\newblock {\em arXiv preprint arXiv:2107.07259}, 2021.

\bibitem{lawrence2021project}
Jason Lawrence, Dan~B Goldman, Supreeth Achar, Gregory~Major Blascovich,
  Joseph~G Desloge, Tommy Fortes, Eric~M Gomez, Sascha H{\"a}berling, Hugues
  Hoppe, Andy Huibers, et~al.
\newblock Project starline: A high-fidelity telepresence system.
\newblock 2021.

\bibitem{li2020monocular}
Ruilong Li, Yuliang Xiu, Shunsuke Saito, Zeng Huang, Kyle Olszewski, and Hao
  Li.
\newblock Monocular real-time volumetric performance capture.
\newblock In {\em European Conference on Computer Vision}, pages 49--67.
  Springer, 2020.

\bibitem{li2019deep}
Yue Li, Pablo Wiedemann, and Kenny Mitchell.
\newblock Deep precomputed radiance transfer for deformable objects.
\newblock {\em Proceedings of the ACM on Computer Graphics and Interactive
  Techniques}, 2(1):1--16, 2019.

\bibitem{lin2022efficient}
Haotong Lin, Sida Peng, Zhen Xu, Yunzhi Yan, Qing Shuai, Hujun Bao, and Xiaowei
  Zhou.
\newblock Efficient neural radiance fields for interactive free-viewpoint
  video.
\newblock In {\em SIGGRAPH Asia 2022 Conference Papers}, pages 1--9, 2022.

\bibitem{lin2022RVM}
Shanchuan Lin, Linjie Yang, Imran Saleemi, and Soumyadip Sengupta.
\newblock Robust high-resolution video matting with temporal guidance.
\newblock In {\em Proceedings of the IEEE/CVF Winter Conference on Applications
  of Computer Vision}, pages 238--247, 2022.

\bibitem{loper2015smpl}
Matthew Loper, Naureen Mahmood, Javier Romero, Gerard Pons-Moll, and Michael~J
  Black.
\newblock Smpl: A skinned multi-person linear model.
\newblock {\em ACM transactions on graphics (TOG)}, 34(6):1--16, 2015.

\bibitem{lorensen1987marching}
William~E Lorensen and Harvey~E Cline.
\newblock Marching cubes: A high resolution 3d surface construction algorithm.
\newblock {\em ACM siggraph computer graphics}, 21(4):163--169, 1987.

\bibitem{lyu2022neural}
Linjie Lyu, Ayush Tewari, Thomas Leimk{\"u}hler, Marc Habermann, and Christian
  Theobalt.
\newblock Neural radiance transfer fields for relightable novel-view synthesis
  with global illumination.
\newblock In {\em Computer Vision--ECCV 2022: 17th European Conference, Tel
  Aviv, Israel, October 23--27, 2022, Proceedings, Part XVII}, pages 153--169.
  Springer, 2022.

\bibitem{ma2021pixel}
Shugao Ma, Tomas Simon, Jason Saragih, Dawei Wang, Yuecheng Li, Fernando
  De~La~Torre, and Yaser Sheikh.
\newblock Pixel codec avatars.
\newblock In {\em Proceedings of the IEEE/CVF Conference on Computer Vision and
  Pattern Recognition}, pages 64--73, 2021.

\bibitem{mescheder2019occupancy}
Lars Mescheder, Michael Oechsle, Michael Niemeyer, Sebastian Nowozin, and
  Andreas Geiger.
\newblock Occupancy networks: Learning 3d reconstruction in function space.
\newblock In {\em Proceedings of the IEEE/CVF conference on computer vision and
  pattern recognition}, pages 4460--4470, 2019.

\bibitem{michalkiewicz2019deep}
Mateusz Michalkiewicz, Jhony~K Pontes, Dominic Jack, Mahsa Baktashmotlagh, and
  Anders Eriksson.
\newblock Deep level sets: Implicit surface representations for 3d shape
  inference.
\newblock {\em arXiv preprint arXiv:1901.06802}, 2019.

\bibitem{mildenhall2021nerf}
Ben Mildenhall, Pratul~P Srinivasan, Matthew Tancik, Jonathan~T Barron, Ravi
  Ramamoorthi, and Ren Ng.
\newblock Nerf: Representing scenes as neural radiance fields for view
  synthesis.
\newblock {\em Communications of the ACM}, 65(1):99--106, 2021.

\bibitem{natsume2019siclope}
Ryota Natsume, Shunsuke Saito, Zeng Huang, Weikai Chen, Chongyang Ma, Hao Li,
  and Shigeo Morishima.
\newblock Siclope: Silhouette-based clothed people.
\newblock In {\em Proceedings of the IEEE/CVF Conference on Computer Vision and
  Pattern Recognition}, pages 4480--4490, 2019.

\bibitem{orts2016holoportation}
Sergio Orts-Escolano, Christoph Rhemann, Sean Fanello, Wayne Chang, Adarsh
  Kowdle, Yury Degtyarev, David Kim, Philip~L Davidson, Sameh Khamis, Mingsong
  Dou, et~al.
\newblock Holoportation: Virtual 3d teleportation in real-time.
\newblock In {\em Proceedings of the 29th annual symposium on user interface
  software and technology}, pages 741--754, 2016.

\bibitem{park2019deepsdf}
Jeong~Joon Park, Peter Florence, Julian Straub, Richard Newcombe, and Steven
  Lovegrove.
\newblock Deepsdf: Learning continuous signed distance functions for shape
  representation.
\newblock In {\em Proceedings of the IEEE/CVF conference on computer vision and
  pattern recognition}, pages 165--174, 2019.

\bibitem{peng2020convolutional}
Songyou Peng, Michael Niemeyer, Lars Mescheder, Marc Pollefeys, and Andreas
  Geiger.
\newblock Convolutional occupancy networks.
\newblock In {\em European Conference on Computer Vision}, pages 523--540.
  Springer, 2020.

\bibitem{peng2021neural}
Sida Peng, Yuanqing Zhang, Yinghao Xu, Qianqian Wang, Qing Shuai, Hujun Bao,
  and Xiaowei Zhou.
\newblock Neural body: Implicit neural representations with structured latent
  codes for novel view synthesis of dynamic humans.
\newblock In {\em Proceedings of the IEEE/CVF Conference on Computer Vision and
  Pattern Recognition}, pages 9054--9063, 2021.

\bibitem{rainer2022neural}
Gilles Rainer, Adrien Bousseau, Tobias Ritschel, and George Drettakis.
\newblock Neural precomputed radiance transfer.
\newblock In {\em Computer Graphics Forum}, volume~41, pages 365--378. Wiley
  Online Library, 2022.

\bibitem{ramamoorthi2001efficient}
Ravi Ramamoorthi and Pat Hanrahan.
\newblock An efficient representation for irradiance environment maps.
\newblock In {\em Proceedings of the 28th annual conference on Computer
  graphics and interactive techniques}, pages 497--500, 2001.

\bibitem{saito2019pifu}
Shunsuke Saito, Zeng Huang, Ryota Natsume, Shigeo Morishima, Angjoo Kanazawa,
  and Hao Li.
\newblock Pifu: Pixel-aligned implicit function for high-resolution clothed
  human digitization.
\newblock In {\em Proceedings of the IEEE/CVF International Conference on
  Computer Vision}, pages 2304--2314, 2019.

\bibitem{saito2020pifuhd}
Shunsuke Saito, Tomas Simon, Jason Saragih, and Hanbyul Joo.
\newblock Pifuhd: Multi-level pixel-aligned implicit function for
  high-resolution 3d human digitization.
\newblock In {\em Proceedings of the IEEE/CVF Conference on Computer Vision and
  Pattern Recognition}, pages 84--93, 2020.

\bibitem{sengupta2018sfsnet}
Soumyadip Sengupta, Angjoo Kanazawa, Carlos~D Castillo, and David~W Jacobs.
\newblock Sfsnet: Learning shape, reflectance and illuminance of facesin the
  wild'.
\newblock In {\em Proceedings of the IEEE conference on computer vision and
  pattern recognition}, pages 6296--6305, 2018.

\bibitem{shao2022doublefield}
Ruizhi Shao, Hongwen Zhang, He Zhang, Mingjia Chen, Yan-Pei Cao, Tao Yu, and
  Yebin Liu.
\newblock Doublefield: Bridging the neural surface and radiance fields for
  high-fidelity human reconstruction and rendering.
\newblock In {\em Proceedings of the IEEE/CVF Conference on Computer Vision and
  Pattern Recognition}, pages 15872--15882, 2022.

\bibitem{shao2022diffustereo}
Ruizhi Shao, Zerong Zheng, Hongwen Zhang, Jingxiang Sun, and Yebin Liu.
\newblock Diffustereo: High quality human reconstruction via diffusion-based
  stereo using sparse cameras.
\newblock {\em arXiv preprint arXiv:2207.08000}, 2022.

\bibitem{sloan2002precomputed}
Peter-Pike Sloan, Jan Kautz, and John Snyder.
\newblock Precomputed radiance transfer for real-time rendering in dynamic,
  low-frequency lighting environments.
\newblock In {\em Proceedings of the 29th annual conference on Computer
  graphics and interactive techniques}, pages 527--536, 2002.

\bibitem{smith2017cyclical}
Leslie~N Smith.
\newblock Cyclical learning rates for training neural networks.
\newblock In {\em 2017 IEEE winter conference on applications of computer
  vision (WACV)}, pages 464--472. IEEE, 2017.

\bibitem{srinivasan2021nerv}
Pratul~P Srinivasan, Boyang Deng, Xiuming Zhang, Matthew Tancik, Ben
  Mildenhall, and Jonathan~T Barron.
\newblock Nerv: Neural reflectance and visibility fields for relighting and
  view synthesis.
\newblock In {\em Proceedings of the IEEE/CVF Conference on Computer Vision and
  Pattern Recognition}, pages 7495--7504, 2021.

\bibitem{sun2019deep}
Ke Sun, Bin Xiao, Dong Liu, and Jingdong Wang.
\newblock Deep high-resolution representation learning for human pose
  estimation.
\newblock In {\em Proceedings of the IEEE/CVF conference on computer vision and
  pattern recognition}, pages 5693--5703, 2019.

\bibitem{suo2021neuralhumanfvv}
Xin Suo, Yuheng Jiang, Pei Lin, Yingliang Zhang, Minye Wu, Kaiwen Guo, and Lan
  Xu.
\newblock Neuralhumanfvv: Real-time neural volumetric human performance
  rendering using rgb cameras.
\newblock In {\em Proceedings of the IEEE/CVF conference on computer vision and
  pattern recognition}, pages 6226--6237, 2021.

\bibitem{tajima2021relightingwild}
Daichi Tajima, Yoshihiro Kanamori, and Yuki Endo.
\newblock Relighting humans in the wild: Monocular full-body human relighting
  with domain adaptation.
\newblock In {\em Computer Graphics Forum}, volume~40, pages 205--216. Wiley
  Online Library, 2021.

\bibitem{tatarchenko2019single}
Maxim Tatarchenko, Stephan~R Richter, Ren{\'e} Ranftl, Zhuwen Li, Vladlen
  Koltun, and Thomas Brox.
\newblock What do single-view 3d reconstruction networks learn?
\newblock In {\em Proceedings of the IEEE/CVF conference on computer vision and
  pattern recognition}, pages 3405--3414, 2019.

\bibitem{tian2022recovering}
Yating Tian, Hongwen Zhang, Yebin Liu, and Limin Wang.
\newblock Recovering 3d human mesh from monocular images: A survey.
\newblock {\em arXiv preprint arXiv:2203.01923}, 2022.

\bibitem{verbin2022ref}
Dor Verbin, Peter Hedman, Ben Mildenhall, Todd Zickler, Jonathan~T Barron, and
  Pratul~P Srinivasan.
\newblock Ref-nerf: Structured view-dependent appearance for neural radiance
  fields.
\newblock In {\em 2022 IEEE/CVF Conference on Computer Vision and Pattern
  Recognition (CVPR)}, pages 5481--5490. IEEE, 2022.

\bibitem{wald2014embree}
Ingo Wald, Sven Woop, Carsten Benthin, Gregory~S Johnson, and Manfred Ernst.
\newblock Embree: a kernel framework for efficient cpu ray tracing.
\newblock {\em ACM Transactions on Graphics (TOG)}, 33(4):1--8, 2014.

\bibitem{wang2004image}
Zhou Wang, Alan~C Bovik, Hamid~R Sheikh, and Eero~P Simoncelli.
\newblock Image quality assessment: from error visibility to structural
  similarity.
\newblock {\em IEEE transactions on image processing}, 13(4):600--612, 2004.

\bibitem{xie2022neural}
Yiheng Xie, Towaki Takikawa, Shunsuke Saito, Or Litany, Shiqin Yan, Numair
  Khan, Federico Tombari, James Tompkin, Vincent Sitzmann, and Srinath Sridhar.
\newblock Neural fields in visual computing and beyond.
\newblock In {\em Computer Graphics Forum}, volume~41, pages 641--676. Wiley
  Online Library, 2022.

\bibitem{hairSeg}
YBIGTA.
\newblock Hair-segmentation.
\newblock \url{https://pytorchhair.gitbook.io/project/}.

\bibitem{yu2021function4d}
Tao Yu, Zerong Zheng, Kaiwen Guo, Pengpeng Liu, Qionghai Dai, and Yebin Liu.
\newblock Function4d: Real-time human volumetric capture from very sparse
  consumer rgbd sensors.
\newblock In {\em Proceedings of the IEEE/CVF Conference on Computer Vision and
  Pattern Recognition}, pages 5746--5756, 2021.

\bibitem{zhang2018unreasonable}
Richard Zhang, Phillip Isola, Alexei~A Efros, Eli Shechtman, and Oliver Wang.
\newblock The unreasonable effectiveness of deep features as a perceptual
  metric.
\newblock In {\em Proceedings of the IEEE conference on computer vision and
  pattern recognition}, pages 586--595, 2018.

\bibitem{zhang2022virtualcube}
Yizhong Zhang, Jiaolong Yang, Zhen Liu, Ruicheng Wang, Guojun Chen, Xin Tong,
  and Baining Guo.
\newblock Virtualcube: An immersive 3d video communication system.
\newblock {\em IEEE Transactions on Visualization and Computer Graphics},
  28(5):2146--2156, 2022.

\bibitem{zheng2021deepmulticap}
Yang Zheng, Ruizhi Shao, Yuxiang Zhang, Tao Yu, Zerong Zheng, Qionghai Dai, and
  Yebin Liu.
\newblock Deepmulticap: Performance capture of multiple characters using sparse
  multiview cameras.
\newblock In {\em Proceedings of the IEEE/CVF International Conference on
  Computer Vision}, pages 6239--6249, 2021.

\bibitem{zheng2021pamir}
Zerong Zheng, Tao Yu, Yebin Liu, and Qionghai Dai.
\newblock Pamir: Parametric model-conditioned implicit representation for
  image-based human reconstruction.
\newblock {\em IEEE transactions on pattern analysis and machine intelligence},
  44(6):3170--3184, 2021.

\bibitem{zheng2019deephuman}
Zerong Zheng, Tao Yu, Yixuan Wei, Qionghai Dai, and Yebin Liu.
\newblock Deephuman: 3d human reconstruction from a single image.
\newblock In {\em Proceedings of the IEEE/CVF International Conference on
  Computer Vision}, pages 7739--7749, 2019.

\end{thebibliography}
}

\end{document}